\documentclass[10pt,twocolumn,letterpaper]{article}
\usepackage{wacv}
\usepackage{times}
\usepackage{epsfig}
\usepackage{graphicx}
\usepackage{amsmath}
\usepackage{amssymb}
\usepackage{tabularx}
\usepackage{booktabs}
\usepackage{soul}
\usepackage{pifont}
\usepackage{makecell}
\newcommand{\cmark}{\ding{51}}
\newcommand{\xmark}{\ding{55}}
\usepackage{subcaption}
\usepackage{multirow}
\usepackage{stackengine}

\newcommand{\figref}[1]{Figure~\ref{#1}}
\newcommand{\tabref}[1]{Table~\ref{#1}}

\usepackage{tabularx}

\graphicspath{{assets/}} 

\usepackage[pagebackref=true,breaklinks=true,letterpaper=true,colorlinks,bookmarks=false]{hyperref}

\wacvfinalcopy 
\def\wacvPaperID{306} 

\def\halfcheckmark{\checkmark\kern-1.2ex\raisebox{.8ex}{\rotatebox[origin=c]{125}{--}} }
\newcolumntype{M}[1]{>{\centering\arraybackslash}m{#1}}
\ifwacvfinal\fi
\begin{document}


\title{The IKEA ASM Dataset: Understanding People Assembling Furniture through Actions, Objects and Pose}

\author{Yizhak Ben-Shabat$^{1,2}$ \qquad Xin Yu$^{3}$ \qquad Fatemeh Sadat Saleh$^{1,2}$ \qquad Dylan Campbell$^{1,2}$\\ Cristian Rodriguez-Opazo$^{1,2}$ \qquad Hongdong Li$^{1,2}$ \qquad Stephen Gould$^{1,2}$\\
\\
${}^{1}$Australian National University 
${}^{2}$Australian Centre for Robotic Vision (ACRV)\\
${}^{3}$University of Technology Sydney \\
\tt\small \url{https://ikeaasm.github.io/}\\
}
\date{}

\maketitle
\ifwacvfinal\thispagestyle{empty}\fi

\begin{figure*}
    \centering
    \includegraphics[width=0.98\textwidth]{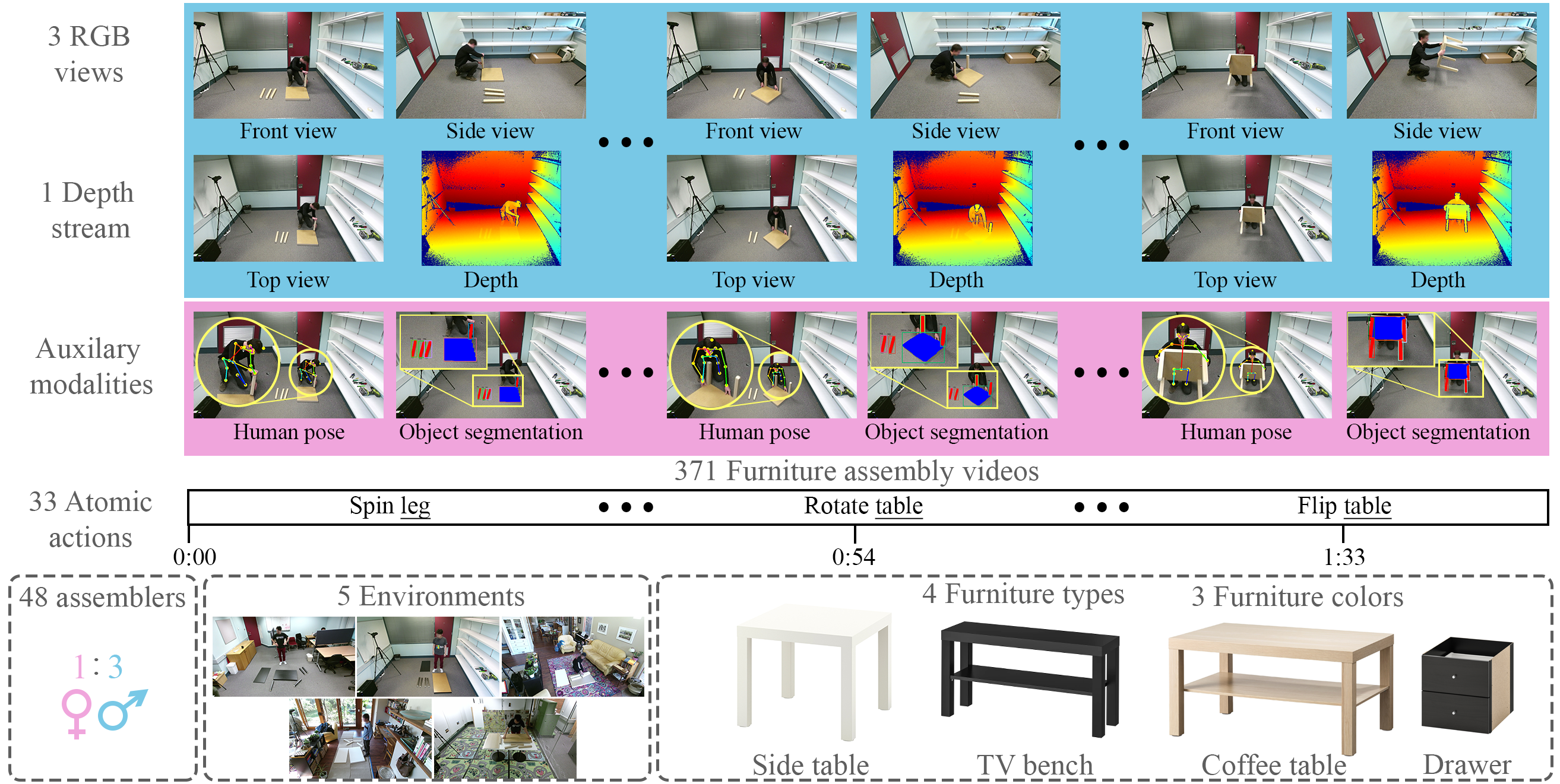}
    \caption{Overview of the IKEA ASM dataset. The dataset contains 371 furniture assembly videos from three camera views, including 3D depth, object segmentation and human pose annotations of 33 atomic actions.}
    \label{fig:teaser}
\end{figure*}

\begin{abstract}
The availability of a large labeled dataset is a key requirement for applying deep learning methods to solve various computer vision tasks. In the context of understanding human activities, existing public datasets, while large in size, are often limited to a single RGB camera and provide only per-frame or per-clip action annotations. To enable richer analysis and understanding of human activities, we introduce IKEA ASM---a three million frame, multi-view, furniture assembly video dataset that includes depth, atomic actions, object segmentation, and human pose. Additionally, we benchmark prominent methods for video action recognition, object segmentation and human pose estimation tasks on this challenging dataset. The dataset enables the development of holistic methods, which integrate multi-modal and multi-view data to better perform on these tasks.
\end{abstract}
\vspace{-3mm}


\section{Introduction}
\label{Sec:intro}

Furniture assembly understanding is closely related to the broader field of action recognition. The rise of deep learning has rapidly advanced this field~\cite{carreira2017quo}. However, deep learning models require vast amounts of training data and are often evaluated on large datasets of short video clips, typically extracted from YouTube~\cite{carreira2017quo, Kuehne11}, that include a set of arbitrary yet highly discriminative actions. Therefore, the research on assembly understanding is far behind generic action recognition due to the insufficient datasets for training such models and other challenges such as the need to understand longer timescale activities. Existing assembly datasets~\cite{toyer2017human} are limited to the classification of very few actions and focus on human pose and color information only. 

We aim to enable research of assembly understanding and underlying perception algorithms under real-life conditions by creating diversity in the assembly environment, assemblers, furniture types and color, and body visibility. To this end, we present the novel IKEA ASM dataset, the first publicly available dataset with the following properties: 
\begin{itemize}
    \itemsep -3pt\partopsep -7pt
    \item \textbf{Multi-modality}: Data is captured from multiple sensor modalities including color, depth, and surface normals. It also includes various semantic modalities including human pose and object instance segmentation. 
    \item \textbf{Multi-view}: Three calibrated camera views cover the work area to handle body, object and self occlusions.
    \item \textbf{Fine-grained}: There is subtle distinction between objects (such as table top and shelf) and action categories (such as aligning, spinning in, and tightening a leg), which are all visually similar.
    \item \textbf{High diversity}: The same furniture type is assembled in numerous ways and over varying time scales. Moreover, human subjects exhibit natural, yet unusual poses, not typically seen in human pose datasets.
    \item \textbf{Transferability}: The straightforward data collection protocol and readily available furniture makes the dataset easy to reproduce worldwide and link to other tasks such as robotic manipulation of the same objects.
\end{itemize}

While the task of furniture assembly is simple and well-defined, there are several difficulties that make inferring actions and detecting relevant objects challenging. First, unlike standard activity recognition the background does not provide any information for classifying the action (since all actions take place in the same environment). Second, parts being assembled are symmetric and highly similar requiring understanding of context and the ability to track objects relative to other parts and sub-assemblies. Third, the strong visual similarity between actions and parts requires a higher-level understanding of the assembly process and state information to be retained over long time periods.

On the other hand, the strong interplay between geometry and semantics in furniture assembly provides an opportunity to model and track the process. Moreover, cues obtained from the different semantic modalities, such as human pose and object types, combine to provide strong evidence for the activity being performed. Our dataset enables research along this direction where both semantics and geometry are important and where short-term feed-forward perception is insufficient to solve the problem.

The main contributions of this paper are: (1) the introduction of a novel furniture assembly dataset that includes multi-view, and multi-modal annotated data; and (2) evaluation of baseline method for different tasks (action recognition, pose estimation, object instance segmentation and tracking) to establish performance benchmarks.

\begin{table*}[]
\setlength\tabcolsep{3pt}
    \centering
    \begin{tabular}{r c c c c c c c c c c c}
        \toprule
         \textbf{Dataset} &  \textbf{Year} & \textbf{Dur.} &\textbf{\#Videos} & \textbf{\#Frames} & \textbf{Activity type} & \textbf{Source}&  \textbf{Views} & \textbf{3D} & \makecell{\textbf{Human} \\ \textbf{pose}} &\makecell{\textbf{object} \\ \textbf{seg. \ bb}} \\
         \hline
         MPII Cooking\cite{rohrbach2012database} & 2012 &9h,28m& 44 & 0.88M& cooking & collected & 1 & \xmark & \cmark&\xmark&\\
         YouCook \cite{das2013thousand} & 2013 &2h,20m& 88& NA & cooking & YouTube & 1 & \xmark & \xmark&\halfcheckmark (bb)&\\
         MPII Cooking 2 \cite{rohrbach2016recognizing} & 2016 & 8h & 273 & 2.88M & cooking & collected & 1 & \xmark & \cmark & \xmark&\\
         IKEA-FA \cite{toyer2017human} & 2017 &3h,50m& 101& 0.41M & assembly & collected & 1 & \xmark & \cmark&\xmark&\\
         YouCook2 \cite{zhou2018towards} & 2018 &176h & 2000 & NA & cooking & YoutTube & 1 & \xmark & \xmark&\halfcheckmark (bb)&\\
         EPIC-Kitchens \cite{damen2018scaling} & 2018 & 55h & 432 & 11.5M & cooking &collected & 1 & \xmark & \xmark&\halfcheckmark (bb)&\\
         COIN \cite{tang2019coin} & 2019 & 476h,38m & 11827 & NA & 180 tasks & YouTube & 1 & \xmark & \xmark& \xmark&\\ 
         Drive\&Act \cite{drive_and_act_2019_iccv} & 2019 &12h &  30 & 9.6M & driving & collected & 6 & \cmark & \cmark & \xmark& \\
         \hline
         IKEA-ASM & 2020 &35h,16m& 371  & 3M & assembly & collected & 3 & \cmark & \cmark& \cmark&\\
          \bottomrule
    \end{tabular}
    \caption{Instructional video dataset comparison.}
    \label{tab:dataset_comp}
\end{table*}

\section{Background and related work}
\label{Sec:related-work}
\noindent \textbf{Related Datasets.}
The increasing popularity of action recognition in the computer vision community has led to the emergence of a wide range of action recognition datasets. 
One of the most prominent datasets for action recognition is the Kinetics~\cite{Qiu_2017_ICCV} dataset---a large-scale human action dataset collected from Youtube videos. It is two orders of magnitude larger than some predecessors, e.g. the UCF101~\cite{soomro2012ucf101} and HMDB51~\cite{kuehne2011hmdb}. Additional notable datasets in this context are ActivityNet~\cite{caba2015activitynet} and  Charades~\cite{sigurdsson2016hollywood}, which include a wide range of human activities in daily life. The aforementioned datasets, while very large in scale, are not domain specific or task-oriented. Additionally, they are mainly centered on single-view RGB data. 

Instructional video datasets usually include domain specific videos, e.g., cooking (MPII~\cite{rohrbach2012database}, YouCook~\cite{das2013thousand}, YouCook2~\cite{zhou2018towards},
 EPIC-Kitchens \cite{damen2018scaling}) and furniture assembly (IKEA-FA~\cite{toyer2017human}). These are most often characterized by having fine grained action labels and may include some additional modalities to the RGB stream such as human pose and object bounding boxes.  There are also more diverse variants like the recent COIN~\cite{tang2019coin} dataset, which forgoes the additional modalities in favor of a larger scale. 
 
 The most closely related to the proposed dataset are the Drive~\&~Act~\cite{drive_and_act_2019_iccv} and NTU RGB+D \cite{shahroudy2016ntu, Liu_2019_NTURGBD120} datasets. Drive~\&~Act is specific to the domain of in-car driver activity and contains multi-view, multi-modal data, including IR streams, pose, depth, and RGB. While the actors follow some instructions, their actions are not task-oriented in the traditional sense. Due to the large effort in collecting it, the total number of videos is relatively low (30). Similarly,  NTU RGB+D \cite{shahroudy2016ntu} and its recent extension NTU RGB+D 120 \cite{Liu_2019_NTURGBD120} contain three different simultaneous RGB views, IR and depth streams as well as 3D skeletons. However, in this case the videos are very short (few seconds), non-instructional and are focused on general activities, some of which are health related or human interaction related. 
 For a detailed quantitative comparison between the proposed and closely-related datasets see Table~\ref{tab:dataset_comp}.

Other notable work is the IKEA Furniture Assembly Environment~\cite{lee2019ikea}, a simulated testbed for studying robotic manipulation. The testbed synthesizes robotic furniture assembly data for imitation learning. Our proposed dataset is complimentary to this work as it captures real-world data of humans that can be used for domain-adaptation. 

In this paper we propose a furniture assembly domain-specific, instructional video dataset with multi-view and multi-modal data, which includes fine grained actions, human pose, object instance segmentation and tracking labels. 

\paragraph{Related methods.} We provide a short summary of methods used as benchmarks in the different dataset tasks including action recognition, instance segmentation, multiple object tracking and human pose estimation. For an extended summary, see the supplementary material. 

\noindent\textbf{Action Recognition .}
Current action recognition architectures for video data are largely image-based. The most prominent approach uses 3D convolutions to extract spatio-temporal features,and includes methods like convolutional 3D (C3D)~\cite{tran2015learning}, which was the first to apply 3D convolutions in this context, pseudo-3D residual net (P3D ResNet)~\cite{Qiu_2017_ICCV}, which leverages pre-trained 2D CNNs and utilizes residual connections and simulates 3D convolutions, and the two-stream inflated 3D ConvNet (I3D)~\cite{carreira2017quo}, which uses an inflated inception module architecture and combines RGB and optical flow streams. Other approaches attempt to decouple visual variations by using a mid-level representation like human pose (skeletons). One idea is to use a spatial temporal graph CNN (ST-GCN)~\cite{yan2018spatial} to process the skeleton's complex structure. Another is to learn skeleton features combined with global co-occurrence patterns~\cite{li2018co}.

\noindent\textbf{Instance Segmentation.}
Early approaches to instance segmentation typically perform segment proposal and classification in two stages~\cite{pinheiro2015learning,dai2016instance,pinheiro2016learning}. Whereas recent one-stage approaches tend to be faster and more accurate~\cite{he2017mask,li2017fully}. Most notably, Mask R-CNN~\cite{he2017mask} combines binary mask prediction with Faster R-CNN~\cite{ren2015faster}, showing impressive performance. They predict segmentation masks on a coarse grid, independent of the instance size and aspect ratio which tends to produce coarse segmentation for instances occupying larger part of the image. To alleviate this problem approaches have been proposed to focus on the boundaries of larger instances, e.g., InstanceCut~\cite{kirillov2017instancecut}, TensorMask~\cite{chen2019tensormask}, and point-based prediction as in PointRend~\cite{kirillov2019pointrend}.  

\noindent\textbf{Multiple Object Tracking (MOT).}
Tracking-by-detection is a common approach for multiple object tracking. MOT can be considered from different aspects: It can be categorized into online or offline, depending on when the decisions are made. In online tracking~\cite{saleh2020artist,wojke2017simple,bergmann2019tracking,chu2019famnet,xu2019spatial,kim2018multi}, the tracker assigns detections to tracklets at every time-step, whereas in offline tracking~\cite{tang2017multiple,maksai2018eliminating} the decision about the tracklets are made after observing the whole video. Different MOT approaches can also be divided into geometry-based~\cite{saleh2020artist,Bewley2016_sort} or appearance-based~\cite{chu2019famnet,bergmann2019tracking,xu2019spatial}. In our context, an application may be human-robot collaboration during furniture assembly, where the tracking system is required to make real-time online decisions~\cite{saleh2020artist,Bewley2016_sort}. In this scenario, IKEA furniture parts are almost textureless and of the same color and shape, and thus the appearance information could be misleading. Additionally, IKEA furniture parts are rigid, non-deformable objects, that are moved almost linearly in a short temporal window. As such, a simple, well-designed tracker that models linear motions~\cite{Bewley2016_sort} is a reasonable choice.

\noindent\textbf{Human Pose Estimation.}
Multi-person 2D pose estimation methods can be divided into bottom-up (predict all joints first)~\cite{pishchulin2016deepcut, cao2017realtime, cao2019openpose, raaj2019efficient} or top-down (detect all person bounding boxes first)~\cite{he2017mask, fang2017rmpe, chen2018cascaded}. The popular OpenPose detector~\cite{cao2017realtime, cao2019openpose} assembles the skeleton using a joint detector and part affinity fields. This was extended to incorporate temporal multi-frame information in Spatio-Temporal Affinity Fields (STAF)~\cite{raaj2019efficient}. Mask R-CNN~\cite{he2017mask} is a notable top-down detection-based approach, where a keypoint regression head can be learned alongside the bounding box and segmentation heads. 
Monocular 3D human pose estimation methods can be categorized as being model-free \cite{pavlakos2018ordinal, pavllo20193d} or model-based \cite{kanazawa2018end, kanazawa2019learning, kolotouros2019convolutional, kocabas2020vibe}. The former include VideoPose3D \cite{pavllo20193d} which estimates 3D joints via temporal convolutions over 2D joint detections in a video sequence. The latter approach predicts the parameters of a body model, often the SMPL model \cite{loper2015smpl}, such as the joint angles, shape parameters, and rotation. Some model-based approaches~\cite{kanazawa2018end, kanazawa2019learning, kocabas2020vibe} leverage adversarial learning to produce realistic body poses and motions. Therefore, they tend to generalize better to unseen datasets, and so we focus on these methods as benchmarks on our dataset.
\section{The IKEA assembly dataset}
The IKEA ASM video dataset will be made publicly available for download of all 371 examples and ground-truth annotations. It includes three RGB views, one depth stream, atomic actions, human poses, object segments, and extrinsic camera calibration. Additionally, we provide code for data processing, including depth to point cloud conversion, surface normal estimation, visualization, and evaluation in a designated github repository. 

\label{Sec:dataset}
\noindent\textbf{Data collection.}
Our data collection hardware system is composed of three Kinect V2 cameras. These three cameras are oriented to collect front, side and top views of the work area. In particular, the top-view camera is set to acquire the scene structure. The front and side-view cameras are placed at eye-level height ($\sim$1.6m). The three Kinect V2 cameras are triggered to capture the assembly activities simultaneously in real time ($\sim$24 fps). To achieve real-time data acquisition performance, multi-threaded processing is used to capture and save images on an Intel i7 8-core CPU with NVIDIA GTX 2080 Ti GPU used for data encoding.

To collect our IKEA ASM dataset, we ask 48 human subjects to assemble furniture in five different environments, such as offices, labs and family homes. 
In this way, the backgrounds are diverse in terms of layout, appearance and lighting conditions. The background is dynamic, containing moving people who are not relevant to the assembly process.
These environments will force algorithms to focus on human action and furniture parts while ignoring the background clutter and other distractors. 
Moreover, to allow human pose diversity, we ask participants to conduct assembly either on the floor or on a table work surface. This yields a total of 10 camera configurations (two per environment).

\begin{figure}
\centering
    \begin{subfigure}{.48\linewidth}
        \centering
    	\includegraphics[width=0.98\linewidth]{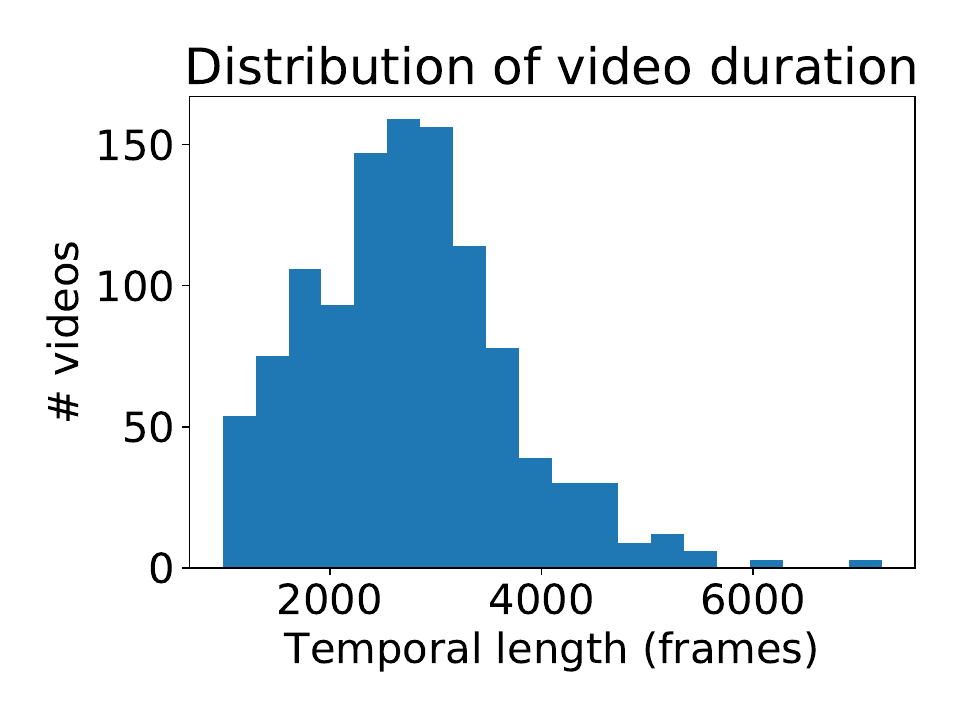}
\end{subfigure}
\begin{subfigure}{.48\linewidth}
    \centering
    \includegraphics[width=0.98\linewidth]{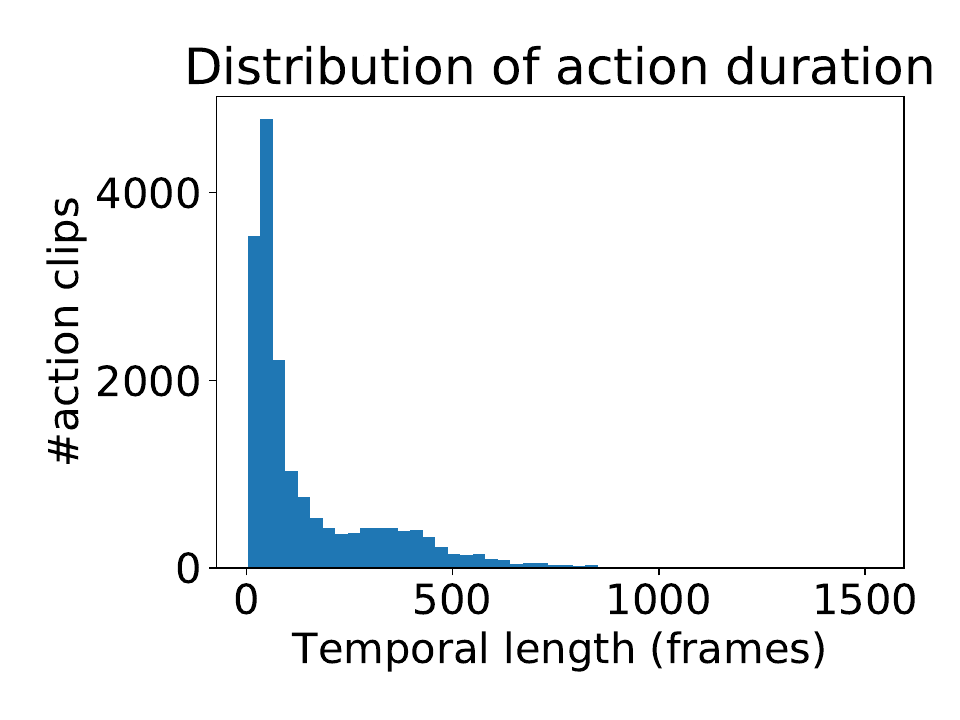}
\end{subfigure}
    \caption{The duration statistics of the videos (left) and actions (right) in the IKEA assembly dataset.}
    \label{fig:video_stats}
\end{figure}

\noindent\textbf{Statistics.} The IKEA ASM dataset consists of 371 unique assemblies  of four different furniture types (side table, coffee table, TV bench, and drawer) in three different colors (white, oak, and black). There are in total 1113 RGB videos and 371 depth videos (top view). Figure~\ref{fig:video_stats} shows the video and individual action length distribution. Overall, the dataset contains 3,046,977 frames ($\sim$35.27h) of footage with an average of 2735.2 frames per video ($\sim$1.89min). 

\begin{figure}
    \centering
    \begin{subfigure}{.49\linewidth}
    \centering
    \includegraphics[width=0.99\linewidth]{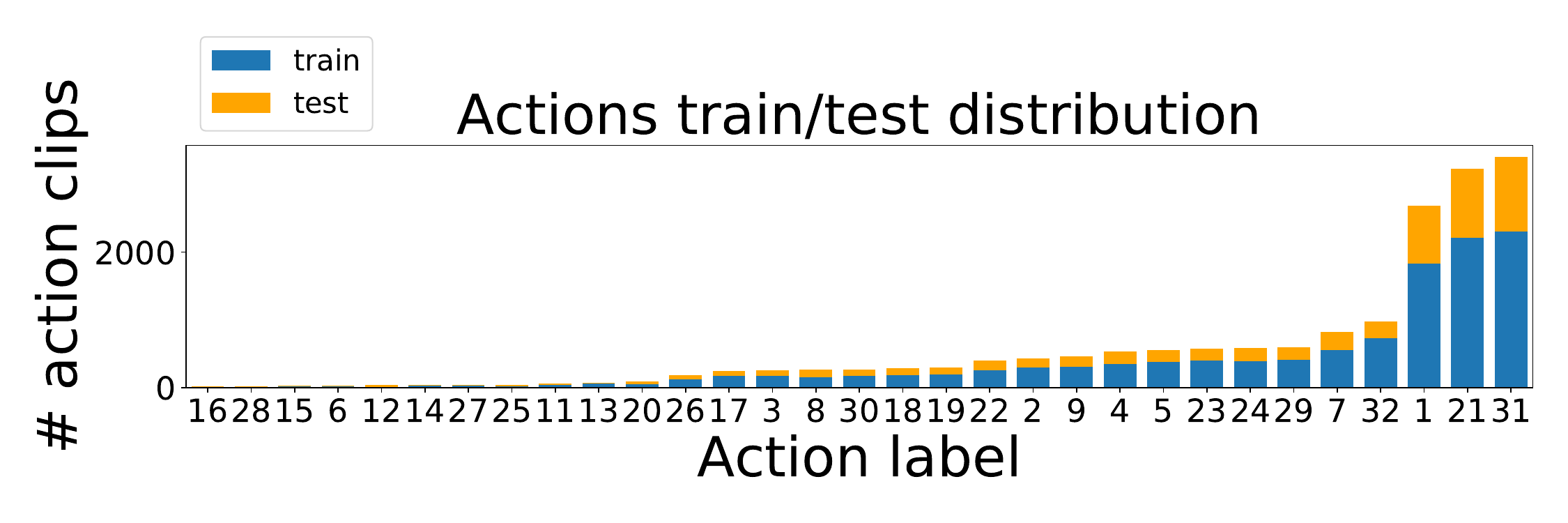}
    \end{subfigure}
        \begin{subfigure}{.49\linewidth}
    \centering
    \includegraphics[width=0.99\linewidth]{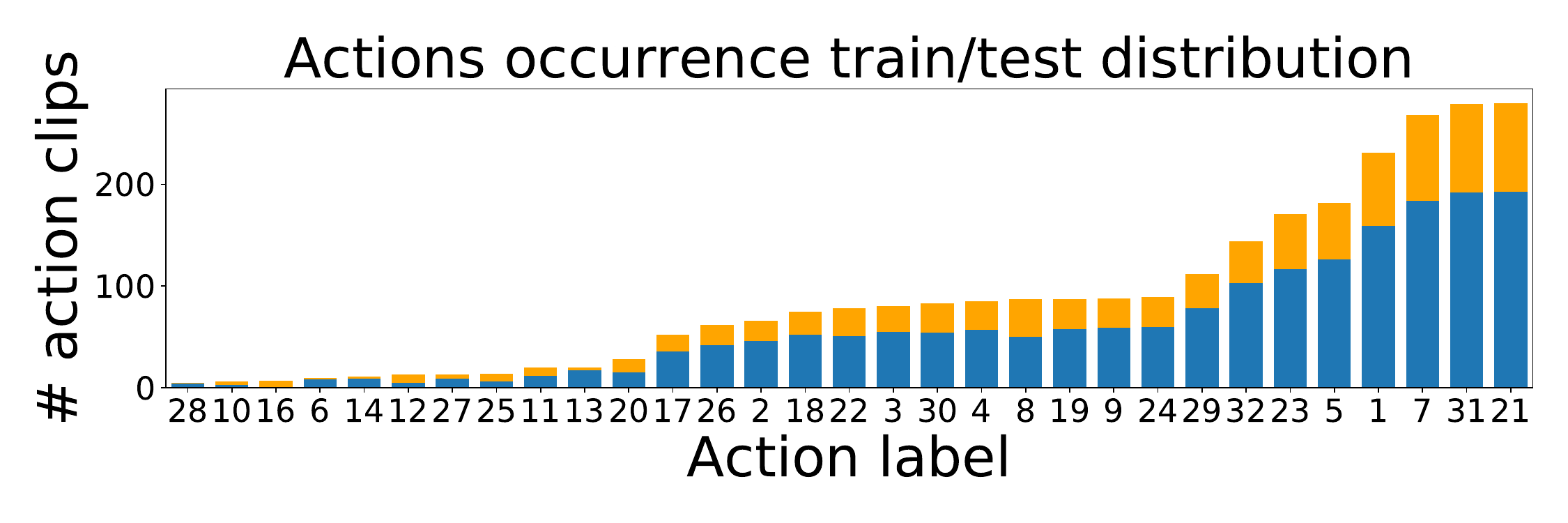}
    \end{subfigure}
    \caption{ The IKEA ASM dataset action distribution (left) and  action instance occurrence distribution (right).}
    \label{fig:action_train_test_split_stats}
\end{figure}

\figref{fig:action_train_test_split_stats} shows the atomic action distribution in the train and test sets.  Each action class contains at least 20 clips. Due to the nature of the assemblies, there is a high imbalance (each table assembly contains four instances of leg assembly). The dataset contains a total of 16,764 annotated actions with an average of 150 frames per action ($\sim$6sec). For a full list of action names and ids, see supplemental.

\noindent\textbf{Data split.}
We aim to enable model training that will generalize to previously unseen environments and human subjects. However, there is a great overlap between subjects in the different scenes and creating a split that will hold-out both simultaneously results in discarding a large portion of the data. Therefore, we propose an environment-based train/test split, i.e., test environments do not appear in the trainset and vise-versa.  The trainset and testset consist of 254 and 117 scans, respectively. Here, test set includes environments 1 and 2 (family room and office). All benchmarks in Section~\ref{Sec:results} were conducted using this split. Additionally, we provide scripts to generate alternative data splits to hold out subjects, environments and joint subject-environments.

\noindent\textbf{Data annotation.}
We annotate our dataset with temporal and spatial information using pre-selected Amazon Mechanical Turk workers to ensure quality. Temporally, we specify the boundaries (start and end frame) of all atomic actions in the video from a pre-defined set. Actions involve interaction with specific object types (e.g., table leg).

Multiple spatial annotations are provided. First, we annotate instance-level segmentation of the objects involved in the assembly. Here an enclosing polygon is drawn around each furniture part. Due to the size of the dataset, we manually annotate only $1\%$ of the video frames which are selected as keyframes that cover diverse object poses and human poses throughout the entire video and provide pseudo ground-truth for the remainder (see \S\ref{sec:instance_segmentation}). Visual inspection was used to confirm the quality of the pseudo ground-truth.
For the same set of manually annotated frames, we also assign each furniture part with a unique ID, which preserves the identity of that part throughout the entire video.

We also annotated the human skeleton of the subjects involved assembly. Here, we asked workers to annotated 12 body joints and five key points related to the face. Due to occlusion with furniture, self-occlusions and uncommon human poses, we include a confidence value between 1 and 3 along with the annotation. Each annotation was then visually inspected and re-worked if deemed to be poor quality.
\section{Experiments and benchmarks}
\label{Sec:results}
We benchmark several state-of-the-art methods for the tasks of frame-wise action recognition, object instance segmentation and tracking, and human pose estimation.

\subsection{Action recognition}
We use three main metrics for evaluation. First, the frame-wise accuracy (FA) which is the de facto standard for action recognition.  We compute it by counting the number of correctly classified frames and divide by the total number of frames in each video and then average over all videos in the test set. Second, since the data is highly imbalanced, we also report the macro-recall by separately computing recall for each category and then averaging. Third, we report the mean average precision (mAP) since all untrimmed videos contain multiple action labels. 
We compare several state-of-the-art methods for action recognition, including I3D~\cite{carreira2017quo}, P3D ResNet~\cite{Qiu_2017_ICCV}, C3D~\cite{tran2015learning}, and frame-wise ResNet~\cite{he2016deep}. For each we start with a pre-trained model and fine-tune it on the IKEA ASM dataset using parameters provided in the original papers. To handle data imbalance we use a weighted random sampler where each class is weighted inversely proportional to its abundance in the dataset.  Results are reported in Table~\ref{tab:results:baseline:action_segmentation} and show that P3D outperforms all other methods, consistent with performance on other datasets. Additionally, the results demonstrate the challenges compared to other datasets where I3D, for example, has an FA score of 57.57\% compared to 68.4\% on Kinetics and 63.64\% on Drive\&Act dataset. 

\begin{table}[]\scriptsize
    \centering
    \begin{tabular}{l c c c c}
         \toprule
            \multirow{2}{*}{\textbf{Method}} &  \multicolumn{2}{c}{\textbf{Frame acc.}}    \\
            & \textbf{top 1} & \textbf{top 3} & \textbf{macro} & \textbf{mAP}\\
            \hline
            ResNet18 \cite{he2016deep} & 27.06 & 55.14 & 21.95 & 11.69\\
            ResNet50 \cite{he2016deep} &  30.38 & 56.1 & 20.03 & 9.47\\
            C3D \cite{tran2015learning} & 45.73 & 69.56  & 32.48 & 21.98\\
            P3D \cite{Qiu_2017_ICCV} & \textbf{60.4} & \textbf{81.07} & \textbf{45.21} & \textbf{29.86}\\
            I3D \cite{carreira2017quo} & 57.57 & 76.55 & 39.34 & 28.59\\
         \bottomrule
    \end{tabular}
    \caption{Action recognition baseline frame-wise accuracy, macro-recall, and mean average precision results.}
    \label{tab:results:baseline:action_segmentation}
\end{table}

\subsection{Multi-view and multi-modal action recognition}
We further explore the affects of multi-view and multi-modal data using the I3D method.
In Table~\ref{tab:results:baseline:action_segmentation_multiview} we report performance on different views and different modalities. We also report their combination by averaging softmax output scores. We clearly see that combining views gives a boost in performance compared to the best single view method. We also find that combining views and pose gives an additional performance increase. Additionally, combining views, depth and pose in the same manner results a small disadvantage, which is due to the inferior performance of the depth based method. This suggests that exploring action recognition in the 3D domain is an open and challenging problem. The results also suggest that a combined, holistic approach that uses multi-view and multi-modal data, facilitated by our dataset, should be further investigated in future work.

\begin{table}[]\scriptsize
\setlength\tabcolsep{3pt}
    \centering
    \begin{tabular}{l l c c c c}
         \toprule
            \multirow{2}{*}{\textbf{Data type}}& \multirow{2}{*}{\textbf{View}} &  \multicolumn{2}{c}{\textbf{Frame acc.}}    \\
                           & & \textbf{top 1} & \textbf{top 3} & \textbf{macro} & \textbf{mAP}\\
            \hline
            \multirow{4}{*}{RGB}&top view   & 57.57 & 76.55 & 39.34 & 28.59 \\ 
            &front view & 60.75 & 79.3 & 42.67 & 32.73\\ 
            &side view  & 52.16 & 72.21 & 36.59 & 26.76\\ 
            &combined views & 63.09 & 80.54 & 45.23 & 32.37 \\
            \hline
            Human pose & HCN \cite{li2018co} & 37.75 & 63.07 & 26.18 & 22.14\\
            Human pose & ST-GCN \cite{yan2018spatial} &  36.99 & 59.63 & 22.77 &
            17.63\\
            &combined RGB+pose & 64.15 & 80.34 & 46.52 & 32.99\\
            \hline
            Depth & top view  & 35.43 & 59.48 & 21.37 & 14.4\\
            \hline
            &combined all & 63.83 & 81.08 & 44.42 & 31.25\\

         \bottomrule
    \end{tabular}
    \caption{Action recognition frame-wise accuracy, macro-recall, and mean average precision results for multi-view/modal inputs.}
    \label{tab:results:baseline:action_segmentation_multiview}
\end{table}

\subsection{Instance segmentation}
\label{sec:instance_segmentation}
As discussed in Section~\ref{Sec:dataset}, the dataset comes with manual instance segmentation annotation for 1\% of the frames (manually selected keyframes that cover diverse object poses and human poses throughout the entire video). To evaluate the performance of existing instance segmentation methods on almost texture-less IKEA furniture, we train Mask R-CNN~\cite{he2017mask} with ResNet50, ResNet101, and ResNeXt101, all with feature pyramid networks structure (FPN) on our dataset.
We train each network using the implementation provided by the Detectron2 framework~\cite{wu2019detectron2}.
Table~\ref{tbl:instance} shows the instance segmentation accuracy for the aforementioned baselines. As expected, the best performing model corresponds to the Mask R-CNN with ResNeXt101-FPN, outperforming ResNet101-FPN and ResNet50-FPN with 3.8\% AP and 7.8\% AP, respectively.

\begin{figure}
    \centering
    \includegraphics[width=0.9\linewidth]{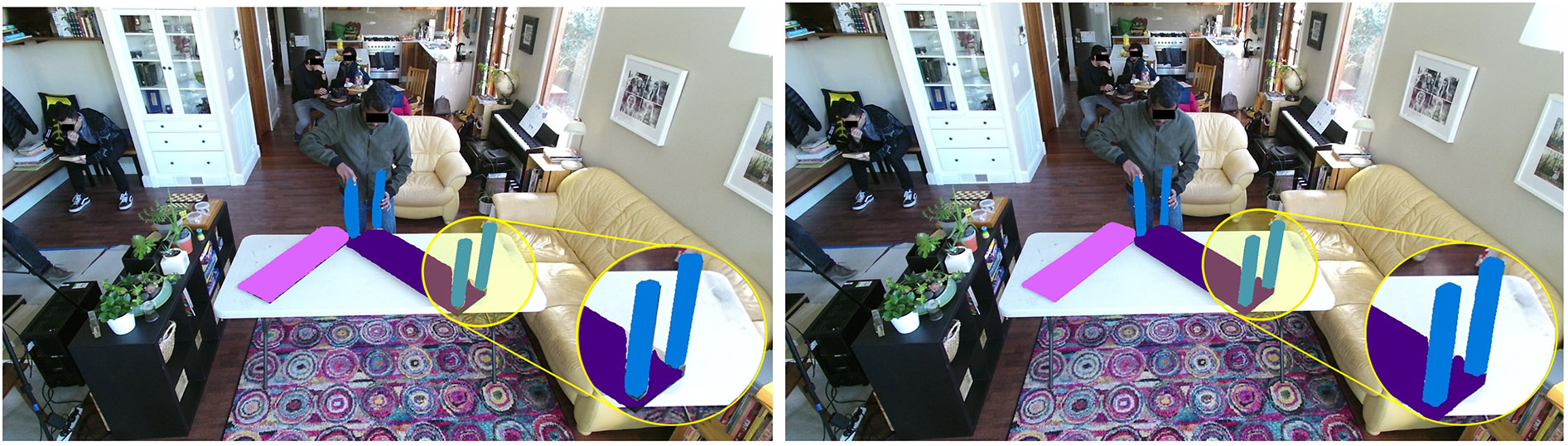}
    \caption{Comparison between pseudo ground-truth without PointRend head (left) and with PointRend head (right).}
    \label{fig:PointVsRes}
\end{figure}

Since the manual annotation only covers 1\% of the whole dataset, we propose to extract pseudo-ground-truth automatically. To this end, we train 12 different Mask R-CNNs with a ResNet50-FPN backbone to overfit on subsets of the training set that cover similar environments and furniture. We show that to achieve manual-like annotations with more accurate part boundaries, training the models with PointRend~\cite{kirillov2019pointrend} as an additional head is essential. Figure~\ref{fig:PointVsRes} compares the automatically generated pseudo-ground-truth with and without the PointRend head. 
To evaluate the effectiveness of adding pseudo-ground-truth, we compare the Mask R-CNN trained with ResNet50-FPN with 1\% annotated data (i.e., manual annotations) and 20\% annotated data (combination of manual and automatically generated annotations) illustrated in Table~\ref{tab:ablation_instance}(a) and see a slight improvement. Note that any backbone architecture can benefit from the automatically generated pseudo-ground-truth.

We also investigate the contribution of adding a PointRend head to the Mask R-CNN with ResNet-50-FPN when training on 1\% of manually annotated data. Table~\ref{tab:ablation_instance}(b) shows that boundary refinement through point-based classification improves the overall instance segmentation performance. This table also clearly shows the effect of PointRend on estimating tighter bounding boxes.

Additionally, to evaluate the effect of furniture color and environment complexity, we report the instance segmentation results partitioned by color (Table~\ref{tab:ablation_instance}(c)) and by environment (Table~\ref{tab:ablation_instance}(d)). Note that, for both of these experiment we use the same model trained on all furniture colors and environments available in the training set. Table~\ref{tab:ablation_instance}(c) shows that oak furniture parts are easier to segment. On the other hand, white furniture parts are the hardest to segment as they reflect the light in the scene more intensely. Another reason is that white parts might be missed due to poor contrast against the white work surfaces.

\begin{figure}
    \centering
    \includegraphics[width=0.95\linewidth]{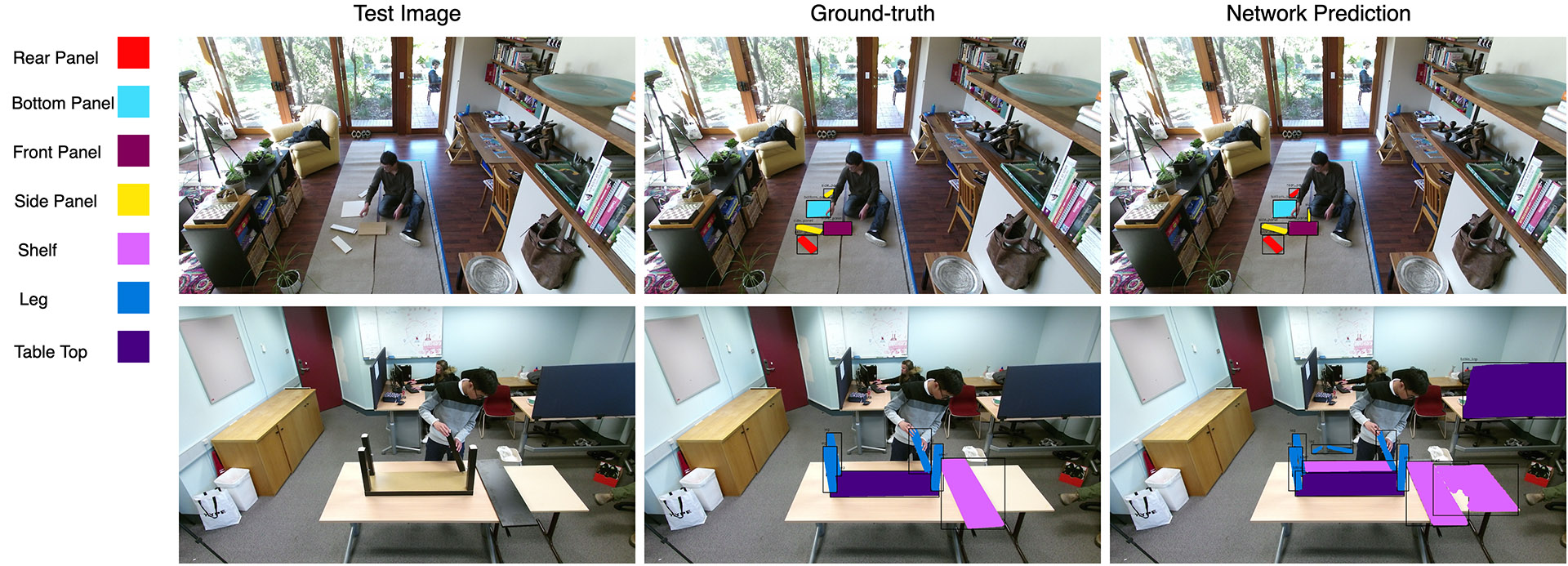}
    \caption{Illustration of part instance segmentation failure cases. (Top row) Mask R-CNN fails to correctly classify different panels of the drawer due to high similarity. (Bottom row) Mask R-CNN incorrectly segments part of the working surface as a furniture part (e.g., shelf) leading to considerable false positives.}
    \label{fig:failures}
\end{figure}

Although Mask R-CNN shows promising results in many scenarios, there are also failure cases, reflecting the real-world challenges introduced by our dataset. These failures are often due to (1) relatively high similarities between different furniture parts, e.g., front panel and rear panel of drawers illustrated in Figure~\ref{fig:failures}(top row) and (2) relatively high similarities between furniture parts of interest and other parts of the environment which introduces false positives. An example of the latter can be seen in Figure~\ref{fig:failures}(bottom row) where Mask R-CNN segmented part of the working surface as the shelf.

\begin{table*}
\scriptsize
\centering
\begin{tabular}{ l c c c c c c c c c c c  } 
\toprule
Feature Extractor & Annotation Type & AP & AP50 & AP75 & table-t & leg & shelf & side-p & front-p & bottom-p & rear-p \\
\midrule
ResNet-50-FPN & mask & 58.1 & 77.2 & 64.2 & 80.8 & 59.8& 68.9 & 32.8 & 50.0 & 66.0 & 48.3\\ 

ResNet-101-FPN & mask & 62.1 & 82.0 & 68.0 & 84.4 & 71.6 & 67.5 & 33.5 & 53.7 & 70.2 & 54.0\\ 
ResNeXt-101-FPN & mask & 65.9 & 85.3 & 73.2 & 87.6 & 71.2 & 76.0 & 44.3 & 52.6 & 73.4 & 56.2\\ 
\midrule
ResNet-50-FPN & bbox & 59.5 & 77.7 & 68.9 & 77.3 & 63.5 & 64.7 & 41.0 & 60.1 & 61.8 & 48.5\\ 

ResNet-101-FPN & bbox & 64.6 & 81.8 & 72.8  & 84.9 & 75.6 & 66.0 & 42.4 & 61.6 & 68.1 & 53.3\\ 
ResNeXt-101-FPN & bbox & 69.5 & 86.4 & 78.9  & 89.4 &  76.8 & 73.7 & 53.3 & 65.8 & 68.7 & 59.0\\ 
\bottomrule
\end{tabular}
\caption{Evaluating the effect of backbone architecture of Mask R-CNN in furniture part instance segmentation.}
\label{tbl:instance}
\end{table*}

\begin{table}[t]
    \centering
\scalebox{0.6}{
    \begin{tabular}{l c c c c c c}
            \multicolumn{6}{c}{\textbf{(a) Influence of adding Pseudo GT}}\\
        \toprule
    Setting & AP$_{segm}$ & AP50$_{segm}$ &AP75$_{segm}$& AP$_{box}$ & AP50$_{box}$ &AP75$_{box}$ \\
    \midrule
     Manual GT & 58.1 & 77.2 & 64.2  & 59.5 & 77.7 & 68.9 \\
    Manual + Pseudo GT & 60.1 & 77.7 & 66.1 & 62.6 & 77.8 & 69.9\\
    \bottomrule
    \\
    \multicolumn{6}{c}{\textbf{(b) Influence of PointRend head}}\\
    \toprule
          Setting & AP$_{segm}$ & AP50$_{segm}$ & AP75$_{segm}$ & AP$_{box}$ & AP50$_{box}$ & AP75$_{box}$\\
         \midrule
         Without PointRend & 58.1 & 77.2 & 64.2  & 59.5 & 77.7 & 68.9 \\
         With PointRend & 61.4 & 80.9 & 67.0 & 63.9 & 82.2 & 73.2 \\
         \bottomrule
         \\
    \multicolumn{6}{c}{\textbf{(c) Color-based Evaluation}}\\
    \toprule
          Colors & AP$_{segm}$ & AP50$_{segm}$& AP75$_{segm}$ & AP$_{box}$ & AP50$_{box}$ & AP75$_{box}$\\
         \midrule
         White & 55.5 & 76.2 & 60.6 & 57.3 & 76.5 & 65.5 \\
         Black & 57.8 & 74.8 & 64.4 & 58.5 & 75.4 & 67.6 \\
         Oak & 62.9 & 82.1 & 69.5 & 64.5 & 82.3 & 75.8\\
         \bottomrule
         \\
         \multicolumn{6}{c}{\textbf{(d) Environment-based Evaluation}}\\
    \toprule
          Environments & AP$_{segm}$ & AP50$_{segm}$ &AP75$_{segm}$& AP$_{box}$ & AP50$_{box}$ & AP75$_{box}$\\
         \midrule
         Env1 (Family Room) & 47.1 & 63.0 & 53.5 & 49.9 & 65.6 & 58.4\\
         Env2 (Office) & 64.4 & 85.0 & 70.7 & 64.8 & 84.3 & 74.6 \\
         \bottomrule
    \end{tabular}
    }

    \caption{Ablation study on furniture part instance segmentation. Note, all experiments are conducted with Mask R-CNN with ResNet-50-FPN as the backbone
    and tested on the same manually annotated data.}
    \label{tab:ablation_instance}
\end{table}{}

\subsection{Multiple furniture part tracking}
\label{sec:tracking}
As motivated in Section~\ref{Sec:dataset}, we utilize SORT~\cite{Bewley2016_sort} as a fast online multiple object tracking algorithm that only relies on geometric information in a class-agnostic manner. Given the detections predicted by the Mask R-CNN, SORT assigns IDs to each detected furniture part at each time-step. 

To evaluate the MOT performance, we use standard metrics~\cite{ristani2016performance,bernardin2008evaluating}.
The main metric is MOTA, which combines three error sources: false positives (FP), false negatives (FN) and identity switches (IDs). A higher MOTA score implies better performance. Another important metric is IDF1, i.e., the ratio of correctly identified detections over the average number of ground-truth and computed detections. The number of identity switches (IDs), FP and FN are also frequently reported. Furthermore, mostly tracked (MT) and mostly lost (ML), that are respectively the ratio of ground-truth trajectories that are covered/lost by the tracker for at least 80\% of their respective life span, provide finer details on the performance of a tracking system. All metrics were computed using the official evaluation code provided by the MOTChallenge benchmark\footnote{\url{https://motchallenge.net/}}.

Table~\ref{tbl:tracking} shows the performance of SORT on each test environment as well as the entire test set. The results reflect the challenges introduced by each environment in the test set. For instance, in Env1 (Family Room) provides a side view of the assembler and thus introduces many occlusions. This can be clearly seen in the number of FN. Moreover, since the tracker may lose an occluded object for a reasonably long time, it may assign new IDs after occlusion, thus affecting the mostly tracked parts and IDF1. On the other hand, the front view provided in Env2 (Office) leads to less occlusions, and thus better identity preservation reflected in IDF1 and MT. However, since the office environment contains irrelevant but similar parts, e.g., the desk partition or the work surface illustrated in Fig.~\ref{fig:failures}(bottom row), we observed considerably higher FP which further affects MOTA.

\begin{table}
\scriptsize
\setlength{\tabcolsep}{3pt}
\begin{tabular}{ l c c c c c c c c} 
\toprule
Test Env. & IDF1$\uparrow$ & MOTA$\uparrow$ & MT$\uparrow$ & PT & ML$\downarrow$ & FP$\downarrow$ & FN$\downarrow$ & IDS$\downarrow$ \\
\midrule
Env1 (Family Room) & 63.7 & 69.6 & 60.1 & 35.9 & 4.0 & 92 & 1152 & 382\\
Env2 (Office)& 72.0 & 59.1 & 94.8 & 5.2 & 0.0 & 4426 & 681 & 370 \\
All & 70.0 & 62.1 & 84.1 & 14.6 & 1.2 & 4518 & 1833 & 752\\
\bottomrule
\end{tabular}
\caption{Evaluating the performance of SORT~\cite{Bewley2016_sort} in multiple furniture part tracking given the detections computed via MASK R-CNN with ResNeXt-101-FPN backbone.}

\label{tbl:tracking}
\end{table}

\subsection{Human pose}
\label{sec:results_human}

\begin{figure}[!t]\centering
\includegraphics[width=0.2\linewidth,
trim=630pt 300pt 500pt 100pt, clip
]{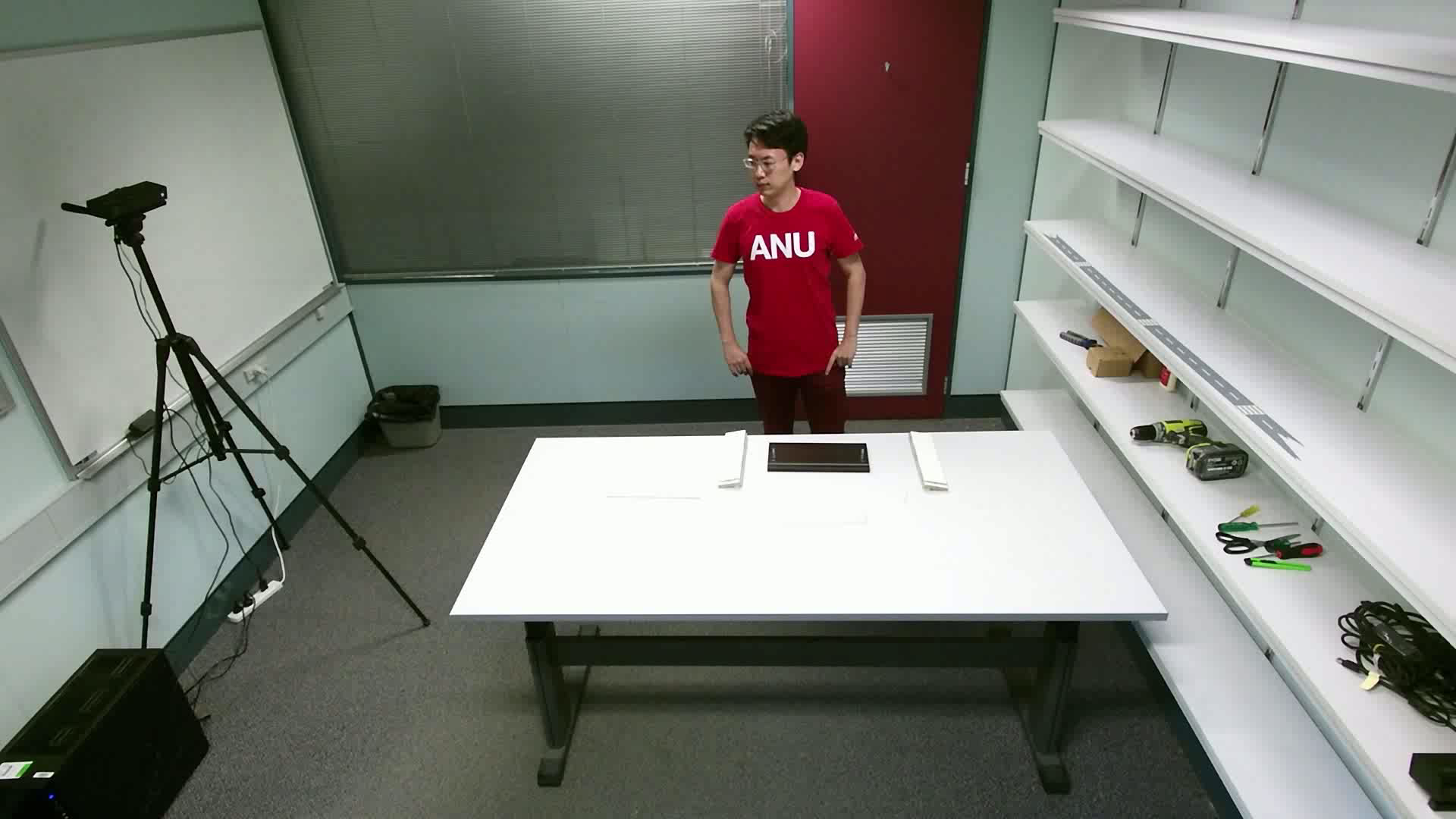}\hfill%
\includegraphics[width=0.2\linewidth,
trim=630pt 300pt 500pt 100pt, clip
]{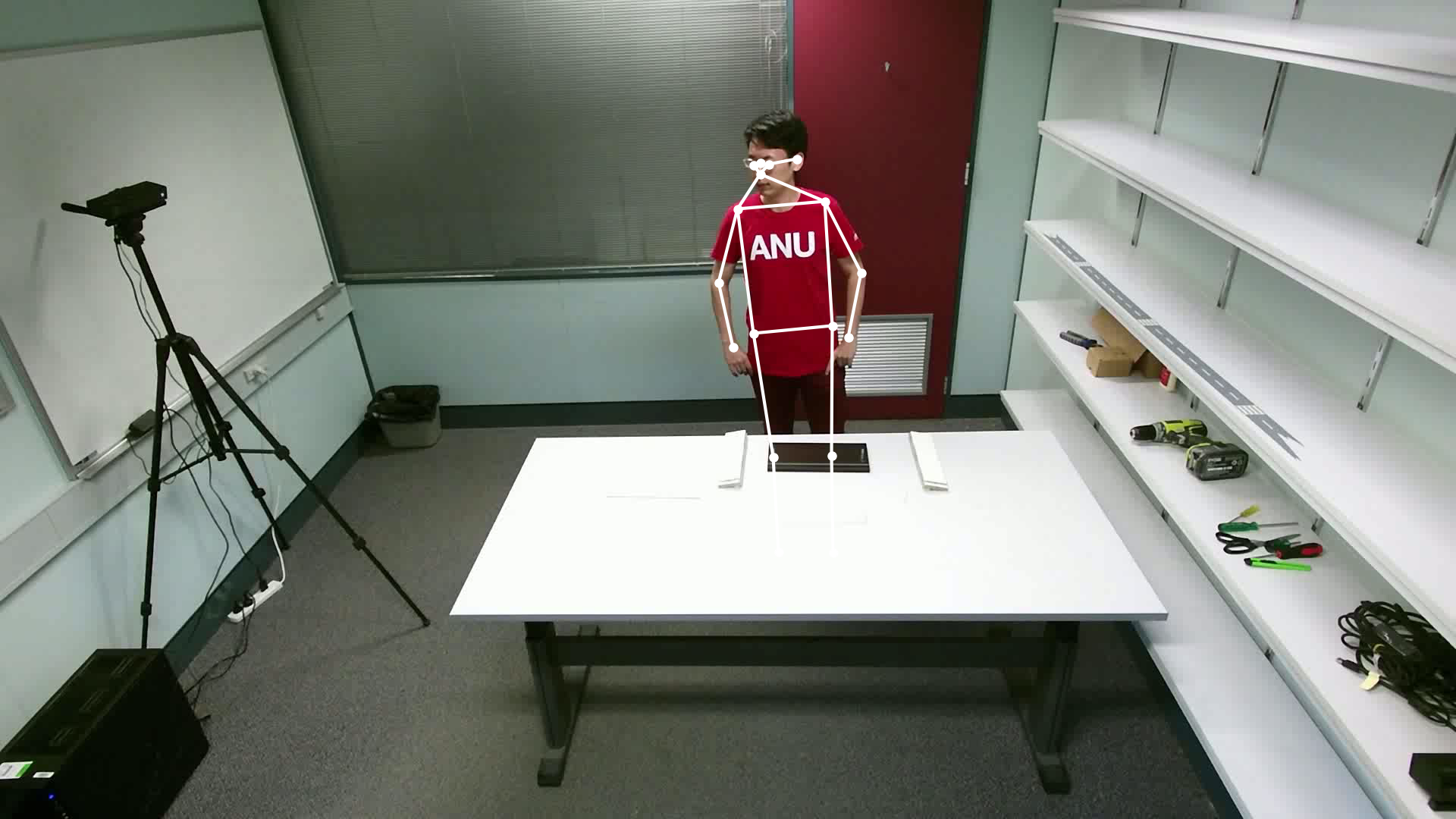}\hfill%
\includegraphics[width=0.2\linewidth,
trim=630pt 300pt 500pt 100pt, clip
]{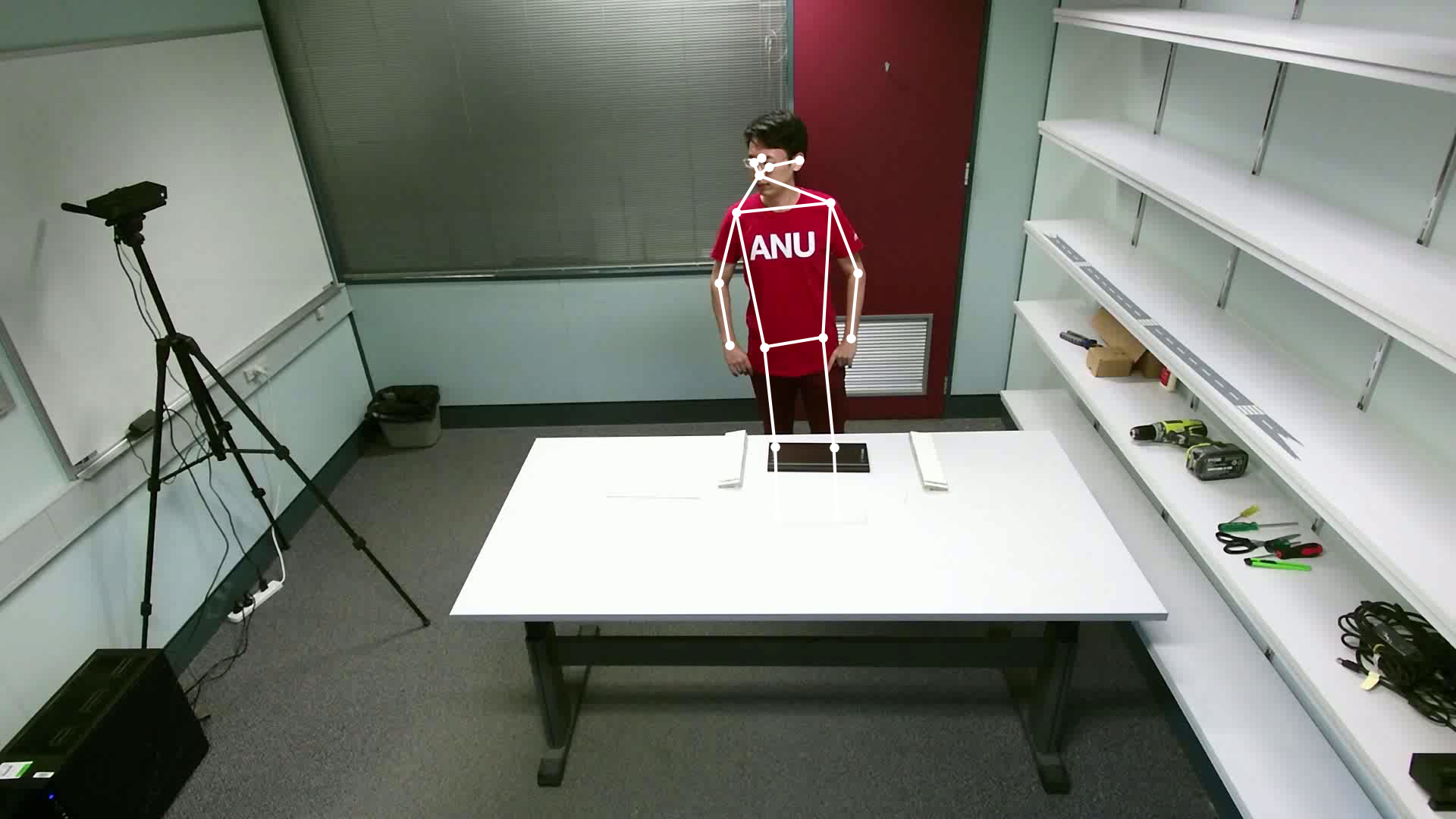}\hfill%
\includegraphics[width=0.2\linewidth, 
trim=100pt 60pt 100pt 60pt, clip
]{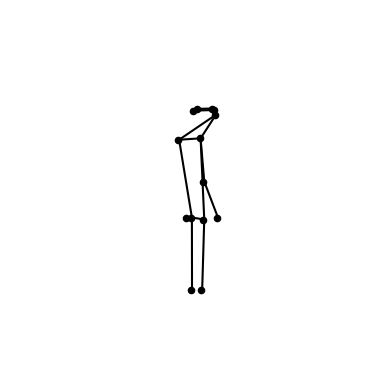}\hfill%
\includegraphics[width=0.2\linewidth, 
trim=100pt 60pt 100pt 60pt, clip
]{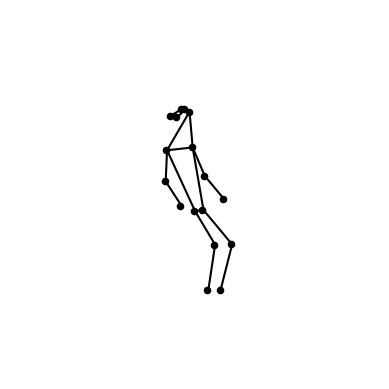}\vspace{0pt}
\includegraphics[width=0.2\linewidth,
trim=260pt 90pt 300pt 150pt, clip
]{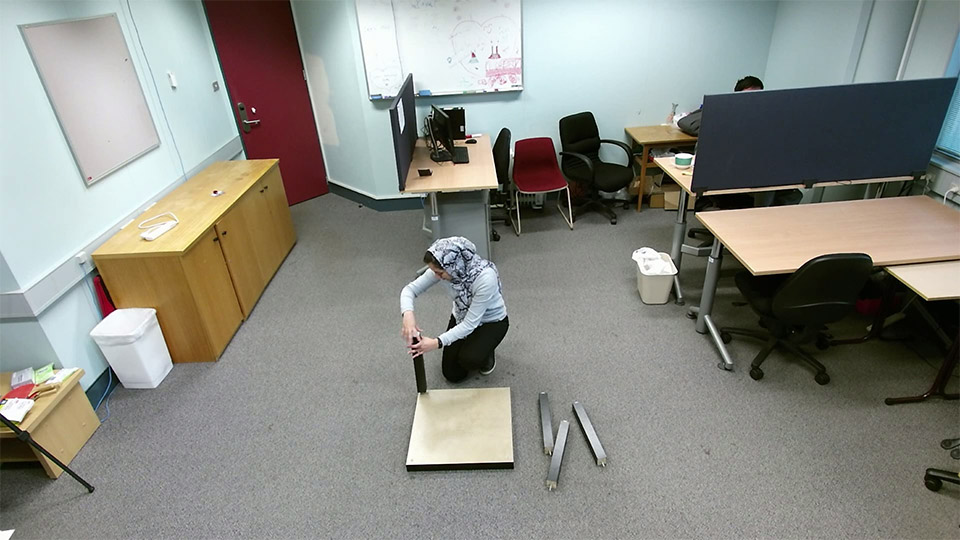}\hfill%
\includegraphics[width=0.2\linewidth,
trim=260pt 90pt 300pt 150pt, clip
]{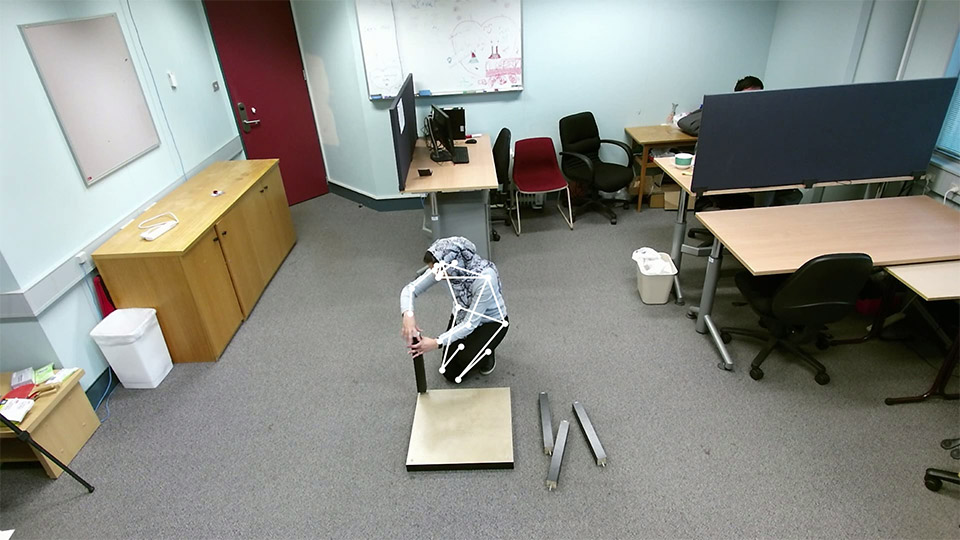}\hfill%
\includegraphics[width=0.2\linewidth,
trim=260pt 90pt 300pt 150pt, clip
]{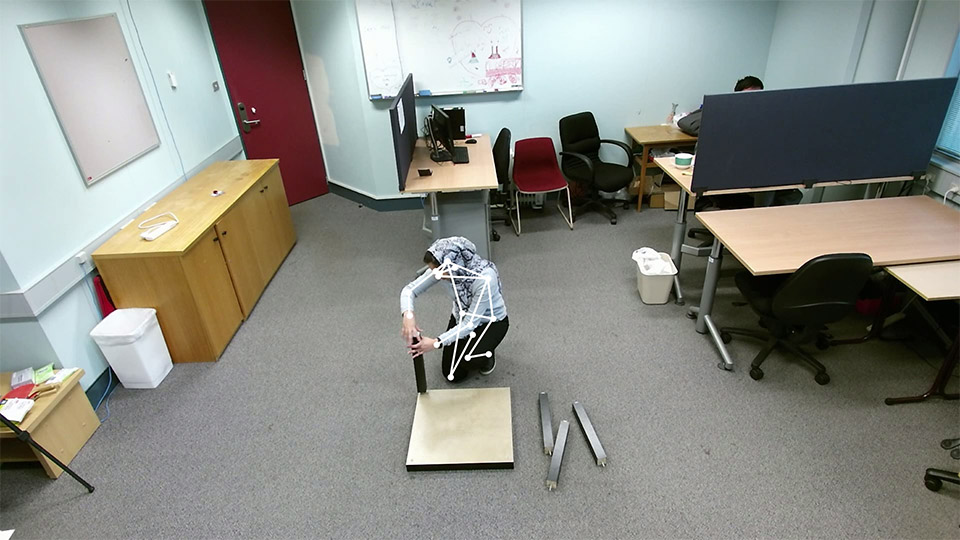}\hfill%
\includegraphics[width=0.2\linewidth, 
trim=100pt 80pt 85pt 80pt, clip
]{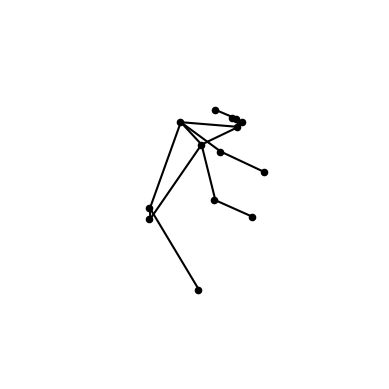}\hfill%
\includegraphics[width=0.2\linewidth, 
trim=100pt 80pt 85pt 80pt, clip
]{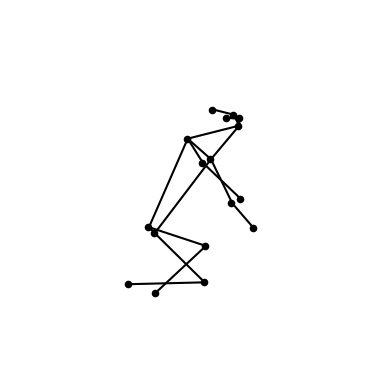}\vspace{0pt}
\includegraphics[width=0.2\linewidth,
trim=275pt 50pt 250pt 195pt, clip
]{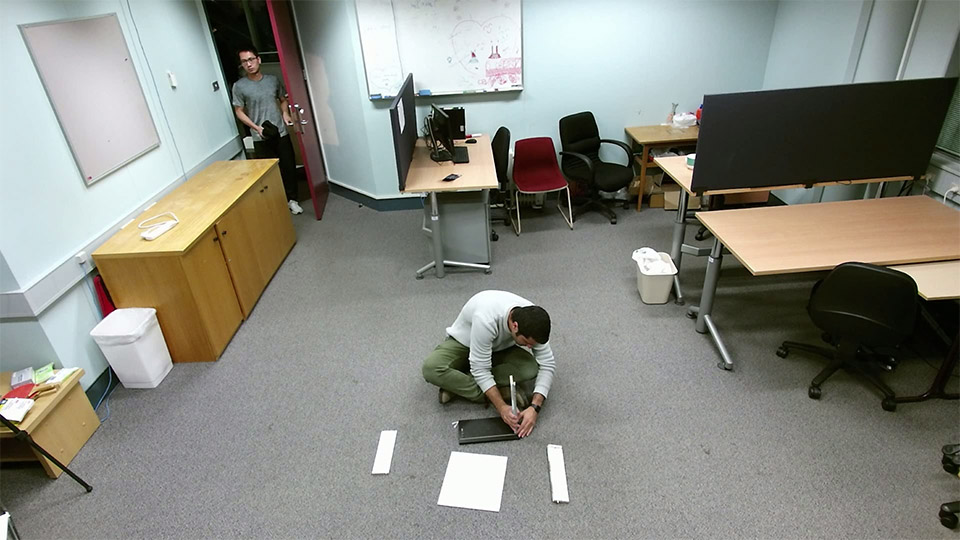}\hfill%
\includegraphics[width=0.2\linewidth,
trim=275pt 50pt 250pt 195pt, clip
]{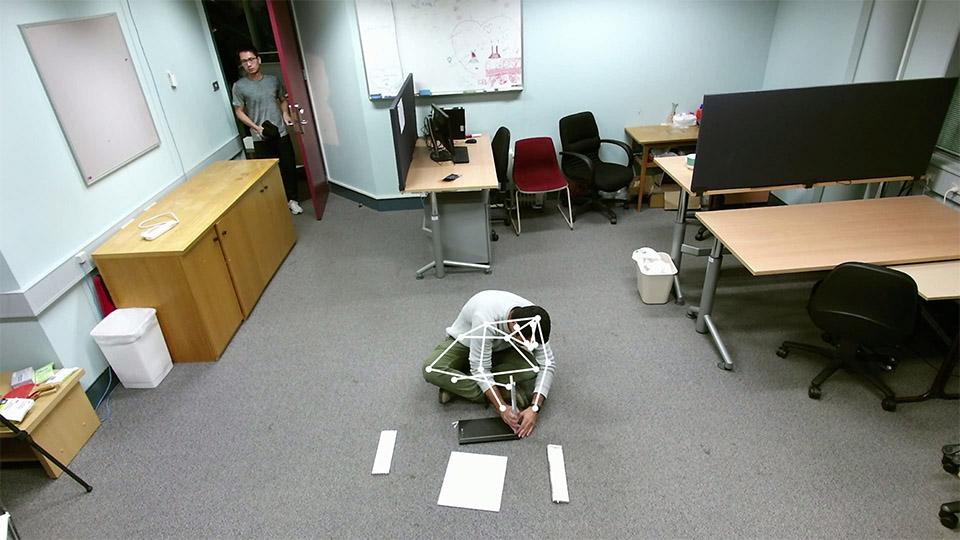}\hfill%
\includegraphics[width=0.2\linewidth,
trim=275pt 50pt 250pt 195pt, clip
]{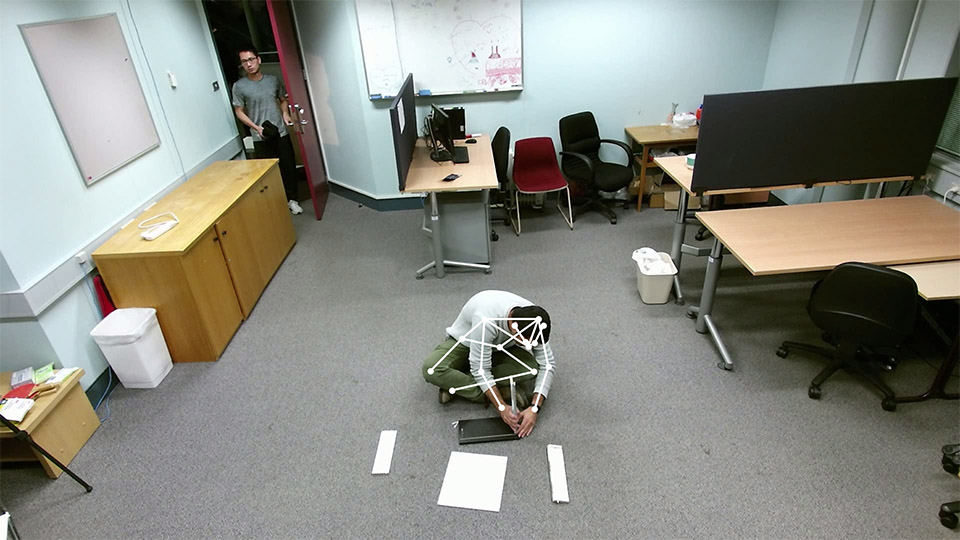}\hfill%
\includegraphics[width=0.2\linewidth, 
trim=100pt 100pt 80pt 110pt, clip
]{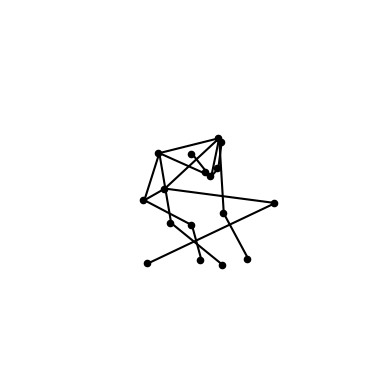}\hfill%
\includegraphics[width=0.2\linewidth, 
trim=100pt 100pt 80pt 110pt, clip
]{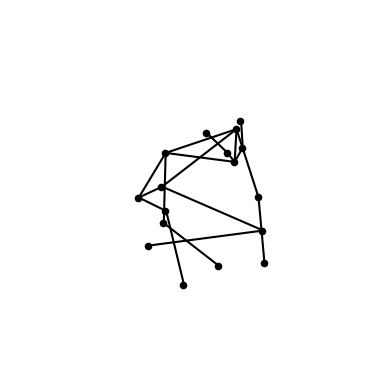}\vspace{0pt}
\caption{Qualitative human pose results. From left to right: sample image, 2D ground-truth, 2D Mask R-CNN prediction, 3D pseudo-ground-truth (novel view), and 3D VIBE prediction (novel view). The middle row shows an example where the 3D pseudo-ground-truth is incomplete, and the bottom row, shows a partial failure case for the predictions.}
\label{fig:human_pose_qualitative_2d}
\end{figure}

The dataset contains 2D human joint annotations in the COCO format \cite{lin2014microsoft} for 1\% of frames, the same keyframes selected for instance segmentation, which cover a diverse range of human poses across each video.
As shown in Figure~\ref{fig:human_pose_qualitative_2d}, there are many highly challenging and unusual poses in the dataset, due to the nature of furniture assembly, particularly when performed on the floor. There are also many other factors that reduce the accuracy of pose estimation approaches, including self-occlusions, occlusions from furniture, baggy clothing, long hair, and human distractors in the background.
We also obtain pseudo-ground-truth 3D annotations by fine-tuning a Mask R-CNN \cite{he2017mask} 2D joint detector on the labeled data, and triangulating the detections of the model from the three calibrated camera views. As a verification step, the 3D points are backprojected to 2D and are discarded if more than 30 pixels from the most confident ground-truth annotations. The reprojection error of the true and pseudo ground-truth annotations is 7.12 pixels on the train set (83\% of ground-truth joints detected) and 9.14 pixels on the test set (53\% of ground-truth joints detected).

To evaluate the performance of benchmark 2D human pose approaches, we perform inference with existing state-of-the-art models, pre-trained by the authors on the large COCO \cite{lin2014microsoft} and MPII \cite{andriluka20142d} datasets and fine-tuned on our annotated data.
We compare OpenPose \cite{cao2017realtime, cao2019openpose}, Mask R-CNN \cite{he2017mask} (with a ResNet-50-FPN backbone \cite{lin2017feature}), and Spatio-Temporal Affinity Fields (STAF) \cite{raaj2019efficient}. The first two operate on images, while the last one operates on videos, and all are multi-person pose estimation methods. We require this since our videos sometimes have multiple people in a frame with only the single assembler annotated.
For fine-tuning, we trained the models for ten epochs with learning rates of 1 and 0.001 for OpenPose and Mask R-CNN, respectively.
We report results with respect to the best detected person per frame, that is, the one that is closest to the ground-truth keypoints, since multiple people may be validly detected in many frames.
We use standard error measures to evaluate the performance of 2D human pose methods: the 2D Mean Per Joint Position Error (MPJPE) in pixels, the Percentage of Correct Keypoints (PCK) \cite{yang2012articulated}, and the Area Under the Curve (AUC) as the PCK threshold varies to a maximum of 100 pixels. A joint is considered correct if it is located within a threshold of 10 pixels from the ground-truth position, which corresponds to 0.5\% of the image width (1080$\times$1920). Absolute measures in pixel space are appropriate for this dataset because the subjects are positioned at an approximately fixed distance from the camera in all scenes. In computing these metrics, only confident ground-truth annotations are used and only detected joints contribute to the mean error (for MPJPE).
The results for 2D human pose baselines on the IKEA ASM train and test sets are reported in Tables~\ref{tab:pose2d}, \ref{tab:pose2d_joints}, and \ref{tab:pose2d_ablation}. 
The best performing model is the fine-tuned Mask R-CNN model, with an MPJPE of 11.5 pixels, a PCK @ 10 pixels of 64.3\% and an AUC of 87.8, revealing considerable room for improvement on this challenging data.
The error analysis shows that upper body joints were detected accurately more often than lower body joints, likely due to the occluding table work surface in half the videos. In addition, female subjects were detected considerably less accurately than male subjects and account for almost all entirely missed detections.

\begin{table}[!t]\centering\scriptsize
\newcolumntype{C}{>{\centering\arraybackslash}X}
\setlength{\tabcolsep}{2pt} 
\begin{tabularx}{\linewidth}{@{}l l C C C | C C C@{}}
\toprule
& & \multicolumn{3}{c|}{Train set} & \multicolumn{3}{c}{Test set}\\
Method & Input & MPJPE$\downarrow$ & PCK$\uparrow$ & AUC$\uparrow$ & MPJPE$\downarrow$ & PCK$\uparrow$ & AUC$\uparrow$\\
\midrule
OpenPose-pt \cite{cao2017realtime} & Image & 17.3 & 46.9 & 78.1 & 16.5 & 46.7 & 77.8\\
OpenPose-ft \cite{cao2017realtime} & Image & 11.8 & 57.8 & 87.7 & 13.9 & 52.6 & 85.6\\ 
MaskRCNN-pt \cite{he2017mask} & Image & 15.5 & 51.9 & 78.2 & 16.1 & 51.5 & 79.2\\ 
MaskRCNN-ft \cite{he2017mask} & Image & \textbf{7.6} & \textbf{77.6} & \textbf{92.1} & \textbf{11.5} & \textbf{64.3} & \textbf{87.8}\\ 
STAF-pt \cite{raaj2019efficient} & Video & 21.4 & 41.8 & 75.3 & 19.7 & 41.1 & 75.4\\
\bottomrule
\end{tabularx}%
\caption{2D human pose results. The Mean Per Joint Position Error (MPJPE) in pixels and the Percentage of Correct Keypoints (PCK) @ 10 pixels ($0.5\%$ image width) are reported.
Pretrained models are denoted `pt' and models fine-tuned on the training data are denoted `ft'.}
\label{tab:pose2d}
\end{table}

\begin{table}[!t]\centering\scriptsize
\newcolumntype{C}{>{\centering\arraybackslash}X}
\setlength{\tabcolsep}{2pt} 
\begin{tabularx}{\linewidth}{@{}l C C C C C C C C@{}}
\toprule
Method & Head & Sho. & Elb. & Wri. & Hip & Knee & Ank. & All\\
\midrule
OpenPose-pt \cite{cao2017realtime} & 65.0 & 34.6 & 59.7 & 68.2 & 12.4 & 29.1 & 24.7 & 46.7\\
OpenPose-ft \cite{cao2017realtime} & 71.0 & 47.9 & 60.4 & 68.8 & 22.7 & 37.7 & 26.3 & 52.6\\ 
MaskRCNN-pt \cite{he2017mask} & 72.5 & 39.0 & 65.8 & 74.1 & 11.3 & 33.9 & 26.1 & 51.5\\ 
MaskRCNN-ft \cite{he2017mask} & 86.5 & 56.4 & 69.4 & 81.9 & 26.9 & 53.4 & 38.0 & 64.3\\
STAF-pt \cite{raaj2019efficient} & 54.9 & 35.1 & 55.9 & 61.0 & 11.6 & 23.4 & 18.7 & 41.1\\
\bottomrule
\end{tabularx}%
\caption{2D human pose test set results per joint group. The Percentage of Correct Keypoints @ 10 pixels is reported.}
\label{tab:pose2d_joints}
\end{table}

\begin{table}[!t]\centering\scriptsize
\newcolumntype{C}{>{\centering\arraybackslash}X}
\setlength{\tabcolsep}{2pt} 
\begin{tabularx}{\linewidth}{@{}l C C c | C C c@{}}
\toprule
& \multicolumn{3}{c|}{Male / Female} & \multicolumn{3}{c}{Floor / Table}\\
Method & MPJPE$\downarrow$ & PCK$\uparrow$ & Miss$\downarrow$ & MPJPE$\downarrow$ & PCK$\uparrow$ & Miss$\downarrow$\\
\midrule
OpenPose-pt \cite{cao2017realtime} & 15.3 / 19.2 & 46.6 / 46.9 & 0 / 1 & 16.9 / 15.8 & 47.1 / 45.9 & 1 / 0\\
OpenPose-ft \cite{cao2017realtime} & 13.8 / 14.0 & 52.4 / 53.0 & 0 / 6 & 14.3 / 13.0 & 52.5 / 52.8 & 0 / 6\\ 
MaskRCNN-pt \cite{he2017mask} & 15.6 / 17.5 & 51.9 / 50.5 & 0 / 1 & 16.8 / 14.8 & 53.1 / 48.3 & 0 / 1\\ 
MaskRCNN-ft \cite{he2017mask} & 11.2 / 11.9 & 64.6 / 63.8 & 0 / 0 & 11.4 / 11.5  & 65.4 / 62.3 & 0 / 0\\
STAF-pt \cite{raaj2019efficient} & 17.6 / 24.1 & 40.7 / 42.1 & 1 / 1 & 19.3 / 20.3 & 39.3 / 44.6 & 1 / 1\\
\bottomrule
\end{tabularx}%
\caption{Evaluating the impact of gender and work surface on 2D human pose test set results.
`Miss' refers to the number of frames in which no joints were detected.
}
\label{tab:pose2d_ablation}
\end{table}

To evaluate the performance of benchmark 3D human pose approaches, we perform inference with existing state-of-the-art models, pre-trained by the authors on large 3D pose datasets,
including Human Mesh and Motion Recovery (HMMR) \cite{kanazawa2019learning}, VideoPose3D (VP3D) \cite{pavllo20193d}, and VIBE \cite{kocabas2020vibe}.
All are video-based methods.
To measure the performance of the different methods, we use the 3D Mean/median Per Joint Position Error (M/mPJPE), which computes the Euclidean distance between the estimated and ground-truth 3D joints in millimeters, averaged over all joints and frames, Procrustes Aligned (PA) M/mPJPE, where the estimated and ground-truth skeletons are rigidly aligned and scaled before evaluation, and the Percentage of Correct Keypoints (PCK) \cite{mehta2017monocular}.
As in the Humans 3.6M dataset \cite{ionescu2014humans36m}, the MPJPE measure is calculated after aligning the centroids of the 3D points in common.
The PCK threshold is set to 150mm, approximately half a head.
The results for 3D human pose baselines on the IKEA ASM dataset are reported in Table~\ref{tab:pose3d}.
The best performing model is VIBE, with a median Procrustes-aligned PJPE of 153mm, and a PA-PCK @ 150mm of 50\%.
%
The baseline methods perform significantly worse on our dataset than standard human pose datasets, demonstrating its difficulty. For example, OpenPose's joint detector \cite{wei2016convolutional} achieves a PCK of 88.5\% on the MPII dataset \cite{andriluka20142d},
compared to 52.6\% on our dataset,
and VIBE has a PA-MPJPE error of 41.4mm on the H36M dataset \cite{ionescu2014humans36m},
compared to 940mm on our dataset.

\begin{table}[!t]\centering\scriptsize
\newcolumntype{C}{>{\centering\arraybackslash}X}
\setlength{\tabcolsep}{1pt} 
\begin{tabularx}{\linewidth}{@{}l l C C C C C C | C C C C C C@{}}
\toprule
& & \multicolumn{6}{c|}{Train set} & \multicolumn{6}{c}{Test set}\\
& & \multicolumn{2}{c}{MPJPE $\downarrow$} & \multicolumn{2}{c}{mPJPE $\downarrow$} & \multicolumn{2}{c|}{PCK $\uparrow$} & \multicolumn{2}{c}{MPJPE $\downarrow$} & \multicolumn{2}{c}{mPJPE $\downarrow$} & \multicolumn{2}{c}{PCK $\uparrow$}\\
Method & PA & \xmark & \cmark & \xmark & \cmark & \xmark & \cmark & \xmark & \cmark & \xmark & \cmark & \xmark & \cmark\\
\midrule
HMMR \cite{kanazawa2019learning} & Vid & 589 & \textbf{501} & 189 & 96 & 32 & 54 & 1012 & 951 & 369 & 196 & 25 & 40\\
VP3D \cite{pavllo20193d} & Vid & \textbf{546} & 518 & \textbf{111} & 87 & \textbf{63} & 70 & \textbf{930} & \textbf{913} & 212 & 179 & \textbf{44} & 47\\
VIBE \cite{kocabas2020vibe} & Vid & 568 & 517 & 139 & \textbf{81} & 55 & \textbf{74} & 963 & 940 & \textbf{199} & \textbf{153} & 43 & \textbf{50}\\
\bottomrule
\end{tabularx}%
\caption{3D human pose results. The Mean Per Joint Position Error (MPJPE) in millimeters, the median PJPE (mPJPE), and the Percentage of Correct Keypoints (PCK) @ 150mm are reported, with and without Procrustes alignment (PA).
Only confident ground-truth annotations are used
and only detected joints contribute to the errors.
}
\label{tab:pose3d}
\end{table}

\section{Conclusion}
\label{Sec:summary}
In this paper, we introduce a large-scale comprehensively labeled furniture assembly dataset for understanding task-oriented human activities with fine-grained actions and common parts. The proposed dataset can also be used as a challenging test-bed for underlying computer vision algorithms such as textureless object segmentation/tracking and human pose estimations in multiple views.
Furthermore, we report benchmark results of strong baseline methods on those tasks for ease of research comparison.
Notably, since our dataset contains multi-view and multi-modal data, it enables the development and analysis of algorithms that use this data, further improving performance on these tasks.
Through recognizing human actions, poses and object positions, we believe this dataset will also facilitate understanding of human-object-interactions and lay the groundwork for the perceptual understanding required for long time-scale structured activities in real-world environments.

\noindent\textbf{Acknowledgements}: This work was supported by The Australian Centre for Robotic Vision and the European Union’s Horizon 2020 research and innovation programme under the Marie Sklodowska-Curie grant agreement No 893465. 

\clearpage
\bibliographystyle{plain}       
\bibliography{references}

\pagebreak
\clearpage
\section{Appendix}
\label{Sec:appendix}
\subsection{Extended related work}
In this section we provide an extended summary of related work for each of the tasks presented in the paper: action recognition, human pose estimation, object instance segmentation and multi-object tracking. 

\noindent \textbf{Action Recognition Methods}:
Current action recognition architectures for video data are largely based on image-based models. These methods employ several strategies for utilizing the additional (temporal) dimension.  One approach is to process the images separately using 2D CNNs and then average the classification results across the temporal domain~\cite{simonyan2014two}. Another approach includes using an RNN instead~\cite{yue2015beyond, donahue2015long}. The most recent and most prominent approach uses 3D convolutions to extract spatio-temporal features, this approach includes the convolutional 3D (C3D) method~\cite{tran2015learning} which was the first to apply 3D convolutions in this context, Pseudo-3D Residual Net (P3D ResNet)~\cite{Qiu_2017_ICCV} which leverages pretrained 2D CNNs,  utilizes residual connections and simulates 3D convolutions, and the two-stream Inflated 3D ConvNet (I3D)~\cite{carreira2017quo} which uses an inflated inception module architecture and combines RGB and optical flow streams. Most recently, the slow-fast method~\cite{feichtenhofer2019slowfast} builds on top of the CNN and processes videos using two frame rates separately to obtain a unified representation.

Another approach for action recognition is to decouple the visual variations and use a mid-level representation like human pose (skeletons). 
Several different approaches were proposed to process the skeleton's complex structure. One approach is to use an LSTM~\cite{liu2016spatio}, another approach is to use a spatial temporal graph CNN (ST-GCN)~\cite{yan2018spatial}. An alternative approach is to encode the skeleton joints and the temporal dynamics in a matrix and process it like an image using a CNN \cite{du2015skeleton, ke2017new}. Similarly, Hierarchical Co-occurrence Network (HCN)~\cite{li2018co}, adopts a CNN to learn skeleton features while leveraging it to learn global co-occurrence patterns.

\paragraph{Instance Segmentation.}
Early approaches to instance segmentation usually combined segment proposal classification in a two-stage framework. For instance, given a number of instance proposal, DeepMask~\cite{pinheiro2015learning} and closely related works~\cite{dai2016instance,pinheiro2016learning} learn to propose instance segment candidates which are then passed through a classifier (e.g., Fast R-CNN). These approaches are usually slower due to the architecture design and tend to be less accurate compared to one-stage counterparts~\cite{he2017mask}. To form a single stage instance segmentation Li et al.~\cite{li2017fully} merged segment proposal and object detection to form a fully convolutional instance segmentation framework. Following this trend, Mask R-CNN~\cite{he2017mask} combines binary mask prediction with Faster R-CNN, showing impressive performance compared to its prior work. 

Mask R-CNN and other similar region-based approaches to instance segmentation~\cite{he2017mask} usually predict segmentation masks on a coarse grid, independent of the instance size and aspect ratio. While this leads to reasonable performance on small objects, around the size of the grid, it tends to produce coarse segmentation for instances occupying larger part of the image. To alleviate the problem of coarse segmentation of large instances, approaches have been proposed to focus on the boundaries of larger instances, e.g., through pixel grouping to form larger masks~\cite{arnab2017pixelwise,liu2017sgn,kirillov2017instancecut} as in InstanceCut~\cite{kirillov2017instancecut}, utilizing sliding windows on the boundaries or complex networks for high-resolution mask prediction as in TensorMask~\cite{chen2019tensormask}, and point-based segmentation prediction as in PointRend~\cite{kirillov2019pointrend}.  

\paragraph{Multiple Object Tracking.}
With the advances in object detection~\cite{redmon2016you,girshick2015fast,ren2015faster}, tracking-by-detection is now a common approach for multiple object tracking (MOT). Mostly studied in the context of multiple person tracking,  MOT can be considered from different aspects. It can be categorized into online or offline, depending on when the decisions are made. In online tracking~\cite{saleh2020artist,wojke2017simple,bergmann2019tracking,chu2019famnet,xu2019spatial,kim2018multi}, the tracker assigns detections to tracklets at every time-step, whereas in offline tracking~\cite{tang2017multiple,maksai2018eliminating} the decision about the tracklets are made after observing the whole context of the video. Different MOT approaches can also be divided into geometry-based~\cite{saleh2020artist,Bewley2016_sort}, appearance-based~\cite{chu2019famnet,bergmann2019tracking,xu2019spatial}, and a combination of appearance and geometry information with social information~\cite{maksai2018eliminating,sadeghian2017tracking}. The choice of information to represent each object highly depends on the context and scenario. For instance, for general multiple person tracking, social information and appearance information could be helpful, but, in sport scenarios, appearance information could be misleading.
In our context for instance, one common application is human-robot collaboration in IKEA furniture assembly, where the tracking system should be able to make its decisions in real-time in an online fashion~\cite{saleh2020artist,Bewley2016_sort}. Moreover, we know that IKEA furniture parts are almost textureless and of the same color and shape, and thus the appearance information could be misleading. Therefore, one may need to employ a completely geometry-based approach. Additionally, we know that IKEA furniture parts are rigid, non-deformable object, that are moved almost linearly in a short temporal window. Therefore, a simple, well-designed MOT that models linear motions~\cite{Bewley2016_sort} is a reasonable choice.

\paragraph{Human Pose Estimation.}
The large volume of work on human pose estimation precludes a comprehensive list; the reader is referred to two recent surveys on 2D and 3D human pose estimation \cite{chen2020monocular, sarafianos20163d} and the references therein. Here, we will briefly discuss recent state-of-the-art approaches, including the baselines selected for our experiments.
Multi-person 2D pose estimation methods can be divided into bottom-up (predict all joints first) \cite{pishchulin2016deepcut, cao2017realtime, cao2019openpose, raaj2019efficient} or top-down (detect all person bounding boxes first) \cite{he2017mask, fang2017rmpe, chen2018cascaded} approaches, with the former reaching real-time processing speeds and the latter having better performance. OpenPose \cite{cao2017realtime, cao2019openpose} uses the CPM joint detector \cite{wei2016convolutional} to predict candidate joint heatmaps and part affinity fields, encoding limb orientation, from which the skeletons can be assembled. This was extended to incorporate temporal multi-frame information in Spatio-Temporal Affinity Fields (STAF) \cite{raaj2019efficient}. Mask R-CNN \cite{he2017mask} is a notable top-down detection-based approach, where a keypoint regression head can be learned alongside the bounding box and segmentation heads. More recently, Cascade Pyramid Networks (CPN) \cite{chen2018cascaded} were proposed, which use multi-scale feature maps and hard keypoint mining to improve position accuracy.
Monocular 3D human pose estimation methods can be categorized as being model-free \cite{pavlakos2018ordinal, pavllo20193d} or model-based \cite{kanazawa2018end, kanazawa2019learning, kolotouros2019convolutional, kocabas2020vibe}. The former include VideoPose3D \cite{pavllo20193d} which estimates 3D joints via temporal convolutions over 2D joint detections in a video sequence. The latter approach predicts the parameters of a body model, often the SMPL model \cite{loper2015smpl}, such as the joint angles, shape parameters, and rotation. For example, Kanazawa \etal \cite{kanazawa2018end} trained an encoder to predict the SMPL parameters, using adversarial learning to encourage realistic body poses and shapes, and later extended this to video input \cite{kanazawa2019learning}. Instead of estimating SMPL parameters, Kolotouros \etal \cite{kolotouros2019convolutional} directly regress the location of mesh vertices using graph convolutions. Finally, Kocabas \etal \cite{kocabas2020vibe} proposed a video-based approach that uses adversarial learning to generate kinematically plausible motions. These model-based approaches tend to generalize better to unseen datasets, and so we focus on these methods as benchmarks on our dataset.

\subsection{Dataset auxiliary data}
In \tabref{tab:appx:taxonomy} we provide the full atmoic action list along with the action identifier, action verb, object name and a short action description. The action class ids correspond to the ids in \figref{fig:action_train_test_split_stats}.
\begin{table}[]
    \centering
    \setlength{\tabcolsep}{3pt}
    \begin{tabular}{c c c c}
         \toprule
            \textbf{ID} & \textbf{Verb}  & \textbf{Object} & \textbf{Description}\\
            \hline
            0 & - &  - & No Annotation (NA) \\
            1 & align & leg & align leg screw with table thread  \\
            \multirow{2}{*}{2} & \multirow{2}{*}{align} & \multirow{2}{*}{side panel} & \multirow{2}{*}{\shortstack{align side panel holes \\ with front panel dowels}}\\
              \\ 
            3 & attach & back panel & attach drawer back panel \\
            4 & attach & side panel &attach drawer side panel\\
            5 & attach & shelf & attach shelf to table \\ 
            6 & flip & shelf & flip shelf \\ 
            7 & flip & table & flip table \\ 
            8 & flip & table top & flip table top \\ 
            9 & insert & pin & insert drawer pin \\
            10 & lay down & back panel & lay down back panel \\ 
            11 & lay down & bottom panel &lay down bottom panel \\
            12 & lay down & front panel & lay down front panel \\
            13 & lay down & leg & lay down leg \\
            14 & lay down & shelf & lay down shelf \\
            15 & lay down & side panel & lay down side panel \\
            16 & lay down & table top & lay down table top \\
            17 &  - & - & other (unavailable action class)\\ 
            18 & pick up & back panel & pick up back panel \\
            19 & pick up & bottom panel & pick up bottom panel \\ 
            20 & pick up & front panel &pick up front panel \\
            21 & pick up & leg & pick up leg \\ 
            22 & pick up & pin & pick up pin \\
            23 & pick up & shelf & pick up shelf \\
            24 & pick up & side panel & pick up side panel \\
            25 & pick up & table top & pick up table top \\
            26 & position & drawer & position the drawer right side up \\
            27 & push & table & push table\\
            28 & push & table top & push table top \\
            29 & rotate & table & rotate table \\
            30 & slide & bottom panel & slide bottom of drawer \\ 
            31 & spin & leg & pin leg \\ 
            32 & tighten & leg & tighten leg \\ 
         \bottomrule
    \end{tabular}
    \caption{Atomic action class id, action verb, object name and action descrtiption.}
    \label{tab:appx:taxonomy}
\end{table}

\subsection{Additional results }
\subsubsection{Action recognition results }
\begin{figure}
    \centering
    \includegraphics[width=0.98\linewidth]{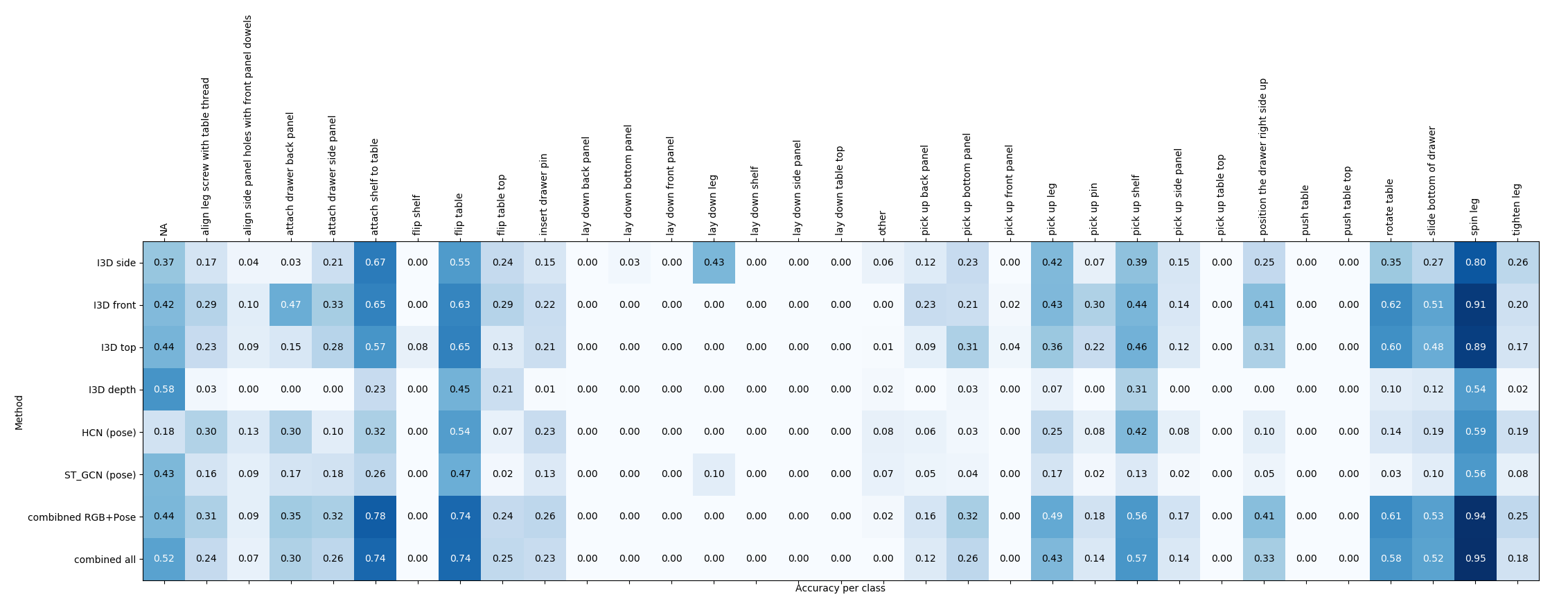}
    \caption{Action recognition accuracy for each class for multi-view/multi-modal baselines.}
    \label{fig:appx:multi-view_results_vis}
\end{figure}

\begin{figure}
    \centering
    \includegraphics[width=0.98\linewidth]{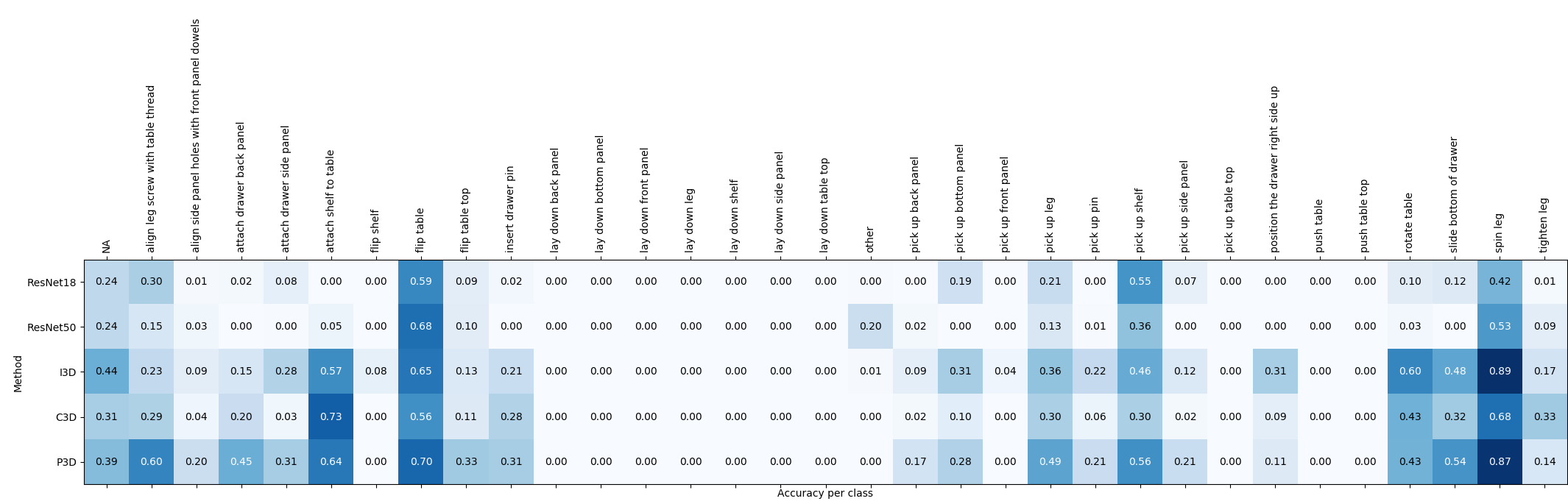}
    \caption{Action recognition accuracy for each class for the baseline methods.}
    \label{fig:appx:per_class_acc_vis}
\end{figure}

We provide additional visualization of the results for the baseline methods and multi-view multi-modal action recognition. In Fig. \ref{fig:appx:multi-view_results_vis} and Fig. \ref{fig:appx:per_class_acc_vis} we visualize the per-class accuracy results of multi-view/multi-modal and action recognition baseline methods respectively. 

\subsubsection{Action localization results }
\begin{table*}[]
    \centering
    \begin{tabular}{l c c c c c c c c c c c c c c}
         \toprule
            \multirow{2}{*}{Method} &  \multicolumn{13}{c}{mAP @ $\alpha$}    \\
                            &0.1 & 0.2 & 0.3 & 0.4 & 0.5 & 0.55 & 0.6 & 0.65 & 0.7 & 0.75 & 0.8 & 0.85 & 0.9 & 0.95 \\
            \hline
            C3D & 0.23 & 0.18 & 0.14 & 0.11 & 0.09 & 0.09 & 0.08 & 0.06 & 0.06 & 0.05 & 0.03 & 0.03 & 0.02 & 0.01   \\
            P3D & 0.38 & 0.34 & 0.28 & 0.23 & 0.17 & 0.15 & 0.13 & 0.11 & 0.09 & 0.08 & 0.06 & 0.05 & 0.02 & 0.01  \\
            I3D & 0.3 & 0.27 & 0.24 & 0.19 & 0.15 & 0.12 & 0.11 & 0.09 & 0.08 & 0.06 & 0.05 & 0.04 & 0.02 & 0    \\
            I3D combined views& 0.44 & 0.39 & 0.33 & 0.27 & 0.2 & 0.18 & 0.16 & 0.13 & 0.12 & 0.1 & 0.09 & 0.06 & 0.03 & 0.01 \\
            I3D combined all & 0.38 & 0.33 & 0.28 & 0.21 & 0.18 & 0.15 & 0.14 & 0.12 & 0.09 & 0.08 & 0.07 & 0.04 & 0.02 & 0.01 \\ 
         \bottomrule
    \end{tabular}
    \caption{Comparison of action localization baselines on the IKEA ASM dataset.}
    \label{tab:results:baseline:action_localization}
\end{table*}
In this section we provide additional baseline results for the task of action localization. The goal in this task is to find and recognize all action instances within an untrimmed test video. The desired output here is a start and end frame for each action appearing in the video sequence. In order to evaluate performance on this task, we follow \cite{caba2015activitynet} and compute the mean average precision (mAP) over all action classes. We set an example as true positive by computing the intersection over union score between the predicted and ground truth temporal segments and checking if it is greater than a threshold $\alpha \in [0.1, 1]$.

\end{document}